\documentclass[letterpaper]{article} 
\usepackage{aaai25}  
\usepackage{times}  
\usepackage{helvet}  
\usepackage{courier}  
\usepackage[hyphens]{url}  
\usepackage{graphicx} 
\usepackage{natbib}  
\usepackage{caption} 
\usepackage{algorithm}
\usepackage{algorithmic}
\usepackage{multirow}
\usepackage{makecell}
\usepackage{bm}
\usepackage{caption}
\usepackage{cite}
\usepackage{amsmath,amssymb,amsfonts}
\usepackage{algorithmic}
\usepackage{textcomp}
\usepackage{xcolor}
\usepackage{subcaption}
\usepackage{pifont}
\def\*#1{\mathbf{#1}}

\usepackage{newfloat}
\usepackage{listings}
\DeclareCaptionStyle{ruled}{labelfont=normalfont,labelsep=colon,strut=off} 
\floatstyle{ruled}
\newfloat{listing}{tb}{lst}{}
\floatname{listing}{Listing}
\title{ITP: Instance-Aware Test Pruning for Out-of-Distribution Detection}
\author{
    Haonan Xu\textsuperscript{\rm 1},
    Yang Yang\textsuperscript{\rm 1,2}\footnote{Corresponding author}
}
\affiliations{
    Nanjing University of Science and Technology\textsuperscript{\rm 1}\\
    Pazhou Lab, Guangzhou\textsuperscript{\rm 2}\\
    \{xhnxhn, yyang\}@njust.edu.cn
}

\usepackage{bibentry}

\begin{document}

\maketitle

\begin{abstract}
Out-of-distribution (OOD) detection is crucial for ensuring the reliable deployment of deep models in real-world scenarios. Recently, from the perspective of over-parameterization, a series of methods leveraging weight sparsification techniques have shown promising performance. These methods typically focus on selecting important parameters for in-distribution (ID) data to reduce the negative impact of redundant parameters on OOD detection. However, we empirically find that these selected parameters may behave overconfidently toward OOD data and hurt OOD detection. To address this issue, we propose a simple yet effective \mbox{post-hoc} method called \mbox{\underline{I}nstance-aware} \underline{T}est \underline{P}runing (\textbf{ITP}), which performs OOD detection by considering both coarse-grained and fine-grained levels of parameter pruning. Specifically, ITP first estimates the class-specific parameter contribution distribution by exploring the ID data. By using the contribution distribution, ITP conducts coarse-grained pruning to eliminate redundant parameters. More importantly, ITP further adopts a fine-grained test pruning process based on the right-tailed \mbox{Z-score} test, which can adaptively remove \mbox{instance-level} overconfident parameters. Finally, ITP derives OOD scores from the pruned model to achieve more reliable predictions. Extensive experiments on widely adopted benchmarks verify the effectiveness of ITP, demonstrating its competitive performance.
\end{abstract}

%

\section{Introduction}
Deep neural networks (DNNs) have recently achieved remarkable success, driving significant progress in various fields, particularly in computer vision \cite{DBLP:conf/iclr/DosovitskiyB0WZ21, DBLP:conf/ijcai/YangZXYZY21, DBLP:conf/aaai/YangHGXX23} and natural language processing \cite{radford2018improving,achiam2023gpt}. However, when deployed in open-world environments, DNNs may fail by producing confident yet erroneous predictions for OOD data. Such unreliable behavior could lead to disastrous consequences, particularly in safety-critical fields like autonomous driving \cite{DBLP:conf/cvpr/GeigerLU12} and medical diagnosis \cite{litjens2017survey}. Therefore, detecting and rejecting predictions on OOD inputs is crucial for ensuring the reliability of AI systems. This task, referred to as OOD detection, has gained widespread attention.

\begin{figure}[t]
    \centering
    \begin{subfigure}{0.8\linewidth}
        \centering
        \includegraphics[width=\linewidth]{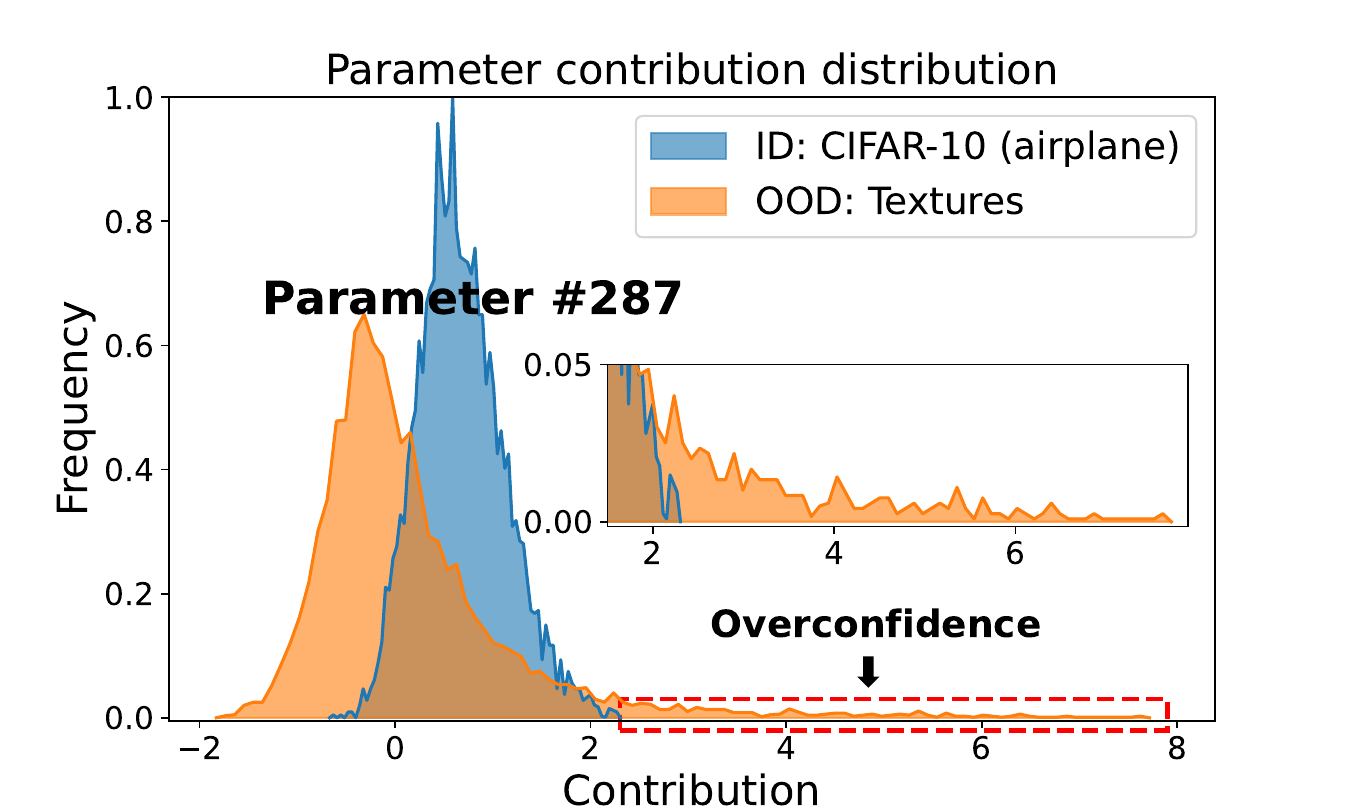}
    \end{subfigure}
    \caption{The distribution of parameter contributions to ID prediction for the CIFAR-10 class ('airplane') on both ID and OOD data. The parameter is selected from the subset of weight parameters that are important for ID prediction in the last layer of DenseNet-101, with pre-ReLU activations utilized for visualization. Since the model outputs are determined by the parameter contribution, overconfident behaviors in the parameters increase the risk of misclassifying OOD data as ID and hurt OOD detection.}
    \label{fig: f1}
\end{figure}

Many techniques have been developed for OOD detection to enhance the discrimination between ID and OOD data. The training-based methods \cite{DBLP:conf/nips/MalininG18,DBLP:conf/iclr/MonteiroRNL023,DBLP:conf/aaai/GhosalSL24} necessitate model training or fine-tuning, whereas post-hoc methods \cite{DBLP:conf/cvpr/BendaleB16,DBLP:conf/icml/SunM0L22,DBLP:conf/cvpr/Wang0F022,DBLP:conf/nips/YangZSX23, DBLP:journals/pami/YangJXZ24} can be applied directly to pre-trained models off-the-shelf, eliminating the need for retraining process. This paper primarily focuses on post-hoc methods, which are easy to use, low-cost, and generally applicable. Early post-hoc methods, such as MSP \cite{DBLP:conf/iclr/HendrycksG17}, Energy \cite{DBLP:conf/nips/LiuWOL20}, and GradNorm \cite{DBLP:conf/nips/HuangGL21}, focus on devising a suitable OOD scoring function based on model outputs or gradients to indicate the likelihood that a sample originates from the OOD distribution. However, these methods overlook the fact that DNNs are typically \mbox{over-parameterized} to fit complex data distributions. This design makes the model more susceptible to noise from redundant parameters \cite{DBLP:conf/eccv/SunL22a}, leading to the brittleness of OOD detection. 

To address this issue, a series of post-hoc methods that use network adjustments to improve OOD scoring has emerged. Sparsification-based methods, represented by DICE \cite{DBLP:conf/eccv/SunL22a} and LINe \cite{DBLP:conf/cvpr/AhnPK23}, provide effective solutions for model over-parameterization. Their key idea is to selectively use the weight parameters that are important for ID data to derive OOD scores, thereby reducing the noise interference caused by redundant parameters in OOD detection. However, as illustrated in Figure \ref{fig: f1}, our empirical findings reveal that these selected important parameters are not always beneficial for OOD detection. When processing OOD data, these parameters may contribute abnormally high to ID predictions, behaving overconfidently. Such overconfidence increases the risk of the model predicting OOD data as ID categories with high confidence. As a result, the derived OOD scores become unreliable, leading to confusion between ID and OOD data and hindering OOD detection. Therefore, addressing parameter-level overconfidence is crucial for better separating ID and OOD data.

Targeting this important problem, we propose Instance-aware Test Pruning (ITP), a simple yet effective method for OOD detection that considers parameter pruning from both coarse-grained and fine-grained perspectives. Concretely, ITP first estimates the class-specific parameter contribution distribution by exploring the ID data. Then, by leveraging the contribution distribution, ITP considers the following two key points to improve OOD detection performance: (1) ITP conducts coarse-grained pruning to remove noise interference caused by redundant parameters based on DICE \cite{DBLP:conf/eccv/SunL22a}. (2) ITP adopts a fine-grained test pruning process based on the right-tailed Z-score test, which adaptively removes instance-level overconfident parameters to reduce the risk of the model making confident yet erroneous predictions. We further provide insightful justification of the working mechanism of ITP from the perspective of the OOD score distribution. As a result of ITP, we show that the OOD scores derived from the model are more reliable and become more separable between ID and OOD data. Moreover, by examining parameter behavior in the weight space, ITP operates orthogonally to activation-based OOD detection methods (e.g., ReAct \cite{DBLP:conf/nips/SunGL21}), facilitating their integration to push ITP's performance further.

\section{Related Work}

\begin{figure*}[htb]
\centering
\includegraphics[width=1.0\textwidth]{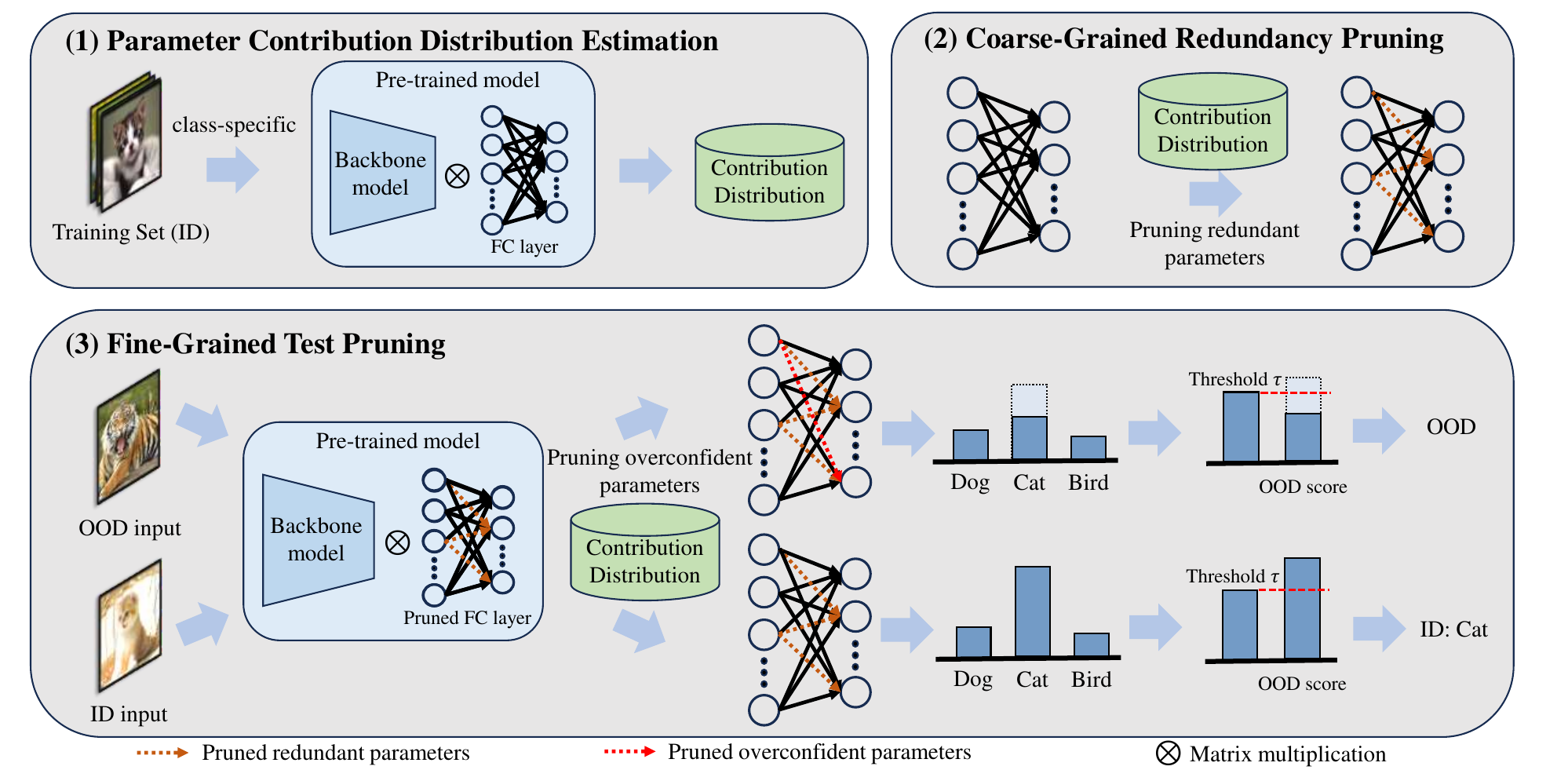}
\caption{Illustration of OOD detection using ITP. The overall procedure involves three main steps. (1) Training data are used to estimate the class-specific parameter contribution distribution for a pre-trained model. (2) Coarse-grained redundancy pruning applies a fixed pruning pattern to the model's last layer to remove redundant parameters. (3) Fine-grained test pruning applies a customized pruning pattern to remove overconfident parameters for each test sample at test time. After applying ITP, the OOD scores derived from the model are better able to distinguish between ID and OOD data.}
\label{fig: model}
\end{figure*}
OOD detection aims to enable models to identify and reject predictions for OOD inputs, thereby ensuring the reliability of AI systems. We highlight three major lines of work.

\noindent\textbf{OOD Scoring Methods} are designed to provide appropriate criteria for indicating the likelihood that an input sample is OOD. Distance-based \cite{DBLP:conf/nips/LeeLLS18,DBLP:conf/aaai/HuangLWCZD21,DBLP:conf/icml/SunM0L22} methods identify OOD data as being farther from the training set compared to ID data. Gradient-based methods \cite{DBLP:conf/nips/HuangGL21,DBLP:conf/nips/BehpourDLHGR23} detect OOD inputs by utilizing information extracted from the gradient space. Output-based methods rely on model output logits to identify OOD data. MSP \cite{DBLP:conf/iclr/HendrycksG17} directly uses the maximum SoftMax score to classify a test sample as either ID or OOD. ODIN \cite{DBLP:conf/iclr/LiangLS18} improves the MSP score by perturbing the input and applying temperature scaling to the logits. Energy score \cite{DBLP:conf/nips/LiuWOL20} uses the logsumexp of the output logits, which is consistent with input density and less susceptible to overconfidence problems. However, output-based methods are often disrupted by redundant or overconfident parameters, which can negatively impact OOD detection.

\noindent\textbf{Sparsification-Based Methods} perform OOD detection by pruning the weights of the model. DICE \cite{DBLP:conf/eccv/SunL22a} proposes selectively using the most salient weights to derive the output for OOD detection. LINe \cite{DBLP:conf/cvpr/AhnPK23} adopts the Shapley value \cite{shapley1953value} for more precise pruning of redundant parameters and neurons, and it further considers the number of activated features by clipping activations. OPNP \cite{DBLP:conf/nips/ChenFLCTY23} prunes the parameters and neurons with exceptionally large or nearly zero sensitivities to mitigate \mbox{over-fitting}. These methods typically focus on selecting parameters that are important for ID prediction before testing for OOD detection. However, at test time, these selected parameters may exhibit overconfidence, which can impact the performance of OOD detection.

\noindent\textbf{Activation-Based Methods} attempt to rectify activations to widen the gap between ID and OOD data. ReAct \cite{DBLP:conf/nips/SunGL21} truncates activations above a pre-computed threshold to treat all activated features equally, thereby incorporating the number of activated features into consideration for OOD detection. VRA \cite{DBLP:conf/nips/XuLLT23} zeros out anomalously low activations and truncates anomalously high activations. BATS \cite{DBLP:conf/nips/ZhuCXLZ00ZC22} proposes rectifying activations towards their typical set, while LAPS \cite{DBLP:conf/aaai/HeYHWSYLG24} improves BATS by considering channel-aware typical sets. These methods only examine anomalies at the activation level, whereas managing overconfidence anomalies at a more granular parameter level is important for more effective OOD detection.

\section{Proposed Method}  

\subsection{Preliminaries}

\noindent\textbf{Setup.} 
In this paper, we follow previous work  \cite{DBLP:conf/nips/YangWZZDPWCLSDZ22} and focus on the setting of $K$-way image classification. Let $\mathcal{X}$ be the input space and $\mathcal{Y} = \left\{1, 2, ..., K\right\}$ be the ID label space. Suppose that the training set $\mathcal{D} = \{(\*x_i, y_i)\}_{i=1}^n$ is drawn \emph{i.i.d} from a joint distribution $\mathcal{P}_{\mathcal{X}\mathcal{Y}}$ defined over $\mathcal{X} \times \mathcal{Y}$. We denote $\mathcal{P}_{in}$ as the marginal distribution of $\mathcal{P}_{\mathcal{X}\mathcal{Y}}$ on $\mathcal{X}$, representing the ID distribution.

Let $f$ be a model pre-trained from $\mathcal{D}$. For typical image classification architectures, $f$ first extracts a $D$-dimensional penultimate feature representation $h(\mathbf{x}) \in \mathbb{R}^D$ from an input $\mathbf{x} \in \mathcal{X}$. The last fully connected (FC) layer, parameterized by a weight matrix $\mathbf{W} \in \mathbb{R}^{D \times K}$ and a bias vector $\mathbf{b} \in \mathbb{R}^K$, then maps $h(\mathbf{x})$ to the output vector $f(\mathbf{x}) \in \mathbb{R}^K$. Mathematically, the model output can be expressed as:
\begin{equation}
f(\mathbf{x}) = \mathbf{W}^\top h(\mathbf{x}) + \mathbf{b}.
\end{equation}

\noindent\textbf{Out-of-distribution Detection.} 
The goal of OOD detection is to determine whether a test input $\*x$ is from $\mathcal{P}_{in}$ (ID) or not (OOD). In practice, the OOD detection task is often formulated as the following binary decision problem: 
\begin{align}
G(\*x) &= 
\begin{cases}
\text{ID},  &  \text{if } S(\*x) > \tau, \\
\text{OOD}, &  \text{if } S(\*x) \leq \tau,
\end{cases}
\end{align}
where $S(\cdot)$ represents the OOD scoring function, and $\tau$ is a chosen threshold to ensure that the majority of ID data are correctly classified (\emph{e.g.}, 95\%). By convention, samples with higher OOD scores are heuristically classified as ID and vice versa. Given that the energy score \cite{DBLP:conf/nips/LiuWOL20} has been proven to be consistent with the input density and performs well, we mainly adopt the negative energy score as the OOD score, expressed as:
\begin{equation}
\label{eq: OOD_score}
S(\*x) = - E\left( {\*x} \right) = \log{\sum\limits_{k = 1}^{K}{\exp({f_{k}(\*x)})}},
\end{equation}
where $E(\*x)$ denotes the energy of $\*x$, and $f_{k}(\*x)$ represents the $k$-th output of the model.

\subsection{Parameter Contribution Distribution Estimation}
\label{sec: pc}
Figure \ref{fig: model} illustrates the overall procedure of our proposal. In this section, we first provide a detailed description of how to estimate the parameter contribution distribution, which will guide subsequent parameter pruning.

\noindent\textbf{Defining the Parameter Contribution.} 
For a given input $\*x$, the contribution of a specific parameter $\bm{\theta}_{ij}$ to the category $k$ is defined as the change in the $k$-th output of the model by the presence or absence (setting $\bm{\theta}_{ij}$ to 0) of the parameter $\bm{\theta}_{ij}$, \emph{i.e.},
\begin{equation}
\label{eq: contribution}
c_{k}(\*x; \bm{\theta}_{ij}) = f_{k}(\*x) - f_{k}(\*x;\bm{\theta}_{ij} = 0).
\end{equation}
Previous studies \cite{DBLP:conf/nips/ZhuCXLZ00ZC22} highlight that the features extracted by early layers show similarities between ID and OOD data. In contrast, the later layers, particularly the penultimate layer, can extract more separable features. In this paper, we primarily focus on the last layer's model parameters, which significantly impact OOD detection by processing separable features and directly influencing particular class outputs. Especially, the contributions of the last layer's parameters $\mathbf{W}_{ij}$ can be expressed more simply using Equation \ref{eq: contribution} as follows (see Appendix for details):
\begin{equation}
\label{eq: pc}
c_{k}(\mathbf{x}; \mathbf{W}_{ij}) =
\begin{cases} 
\mathbf{W}_{ij} \cdot h_{i}(\mathbf{x}), & \text{if } k = j, \\
0, & \text{if } k \ne j.
\end{cases}
\end{equation}

\noindent\textbf{Estimating the Distribution of Parameter Contribution.} 
The distribution estimation relies on the assumption that the contribution of the last layer's parameters approximately follows Gaussian distributions parameterized by $(\mu, \sigma)$, as observed empirically (see Appendix). According to the Equation \ref{eq: pc}, the parameter $\mathbf{W}_{ij}$ is specifically associated with class $j$. To minimize potential bias from including data from other classes, we estimate the contribution distribution of parameter $\mathbf{W}_{ij}$ using only the training data for class $j$. Let $\mathcal{D}_{j}$ denote the set of data points belonging to class $j$. The mean $\mu_{ij}$ and standard deviation $\sigma_{ij}$ of the contribution distribution for the parameter $\mathbf{W}_{ij}$ are estimated using the following class-specific formulas:
\begin{equation}
\begin{aligned}
\mu_{ij} &= \frac{1}{|\mathcal{D}_{j}|} \sum_{x \in \mathcal{D}_{j}} c_{j}(x; \mathbf{W}_{ij}), \\
\sigma_{ij} &= \left(\frac{1}{|\mathcal{D}_{j}| - \delta} \sum_{x \in \mathcal{D}_{j}} \left( c_{j}(x; \mathbf{W}_{ij}) - \mu_{ij} \right)^{2}\right)^{\frac{1}{2}},
\end{aligned}
\end{equation}
where $\lvert \mathcal{D}_{j} \rvert$ denotes the cardinality of the set $\mathcal{D}_{j}$, and $\delta$ represents the correction factor. To correct the bias in the estimation of the population standard deviation, we adopt Bessel's correction by setting $\delta$ to 1.

\subsection{Instance-Aware Test Pruning (ITP)}
\label{sec:33}
In this section, we introduce two parameter pruning strategies with different levels of granularity used in ITP for post-hoc enhancement in OOD detection: coarse-grained redundancy pruning (Figure \ref{fig: model}(2)) and fine-grained test pruning (Figure \ref{fig: model}(3)). Detailed descriptions of each strategy are provided below.

\noindent\textbf{Coarse-Grained Redundancy Pruning (CRP)} remove redundant parameters in the over-parameterized weight space of the model. CRP operates at a coarse-grained level by applying a uniform pruning pattern across all test samples. Specifically, CRP measures each parameter's redundancy based on its average contribution to ID prediction. Parameters that fall within the lowest $p\%$ of average contributions are deemed redundant and pruned. To implement this, we define a mask matrix $\*M^{\text{CRP}}$ for CRP, where the average contribution of a parameter is directly obtained from the mean $\mu$ of its contribution distribution. The $(i, j)$-th entry of the mask matrix $\*M_{ij}^{\text{CRP}} \in \*M^{\text{CRP}}$ is defined as follows:
\begin{equation}
\*M_{ij}^{\text{CRP}} = 
\begin{cases} 
1, & \text{if}\quad \mu_{ij} > \Omega_p, \\
0, & \text{if}\quad \mu_{ij} \leq \Omega_p,
\end{cases}
\end{equation}
where $\Omega_p$ represents the average contribution threshold at the lowest $p$ percentile. The model output after applying CRP can be expressed as follows:
\begin{equation}
f^{\text{CRP}}(\*x) = \left( {\*W \odot \*M^{\text{CRP}}} \right)^{\top}h(\*x) + \*b,
\end{equation}
where $\odot$ denotes the element-wise multiplication. Through CRP, we remove noise interference from redundant parameters at a coarse-grained level in OOD detection by retaining only the parameters important for ID data, thereby enhancing the distinction between ID and OOD data. 

\noindent\textbf{Fine-Grained Test Pruning (FTP)} prunes overconfident parameters with anomalously high contributions to ID prediction. FTP operates at a fine-grained level by customizing parameter pruning patterns for each test sample at test time. Specifically, FTP determines whether the parameter is overconfident by performing a right-tail test based on the Z-score. The Z-score quantifies the deviation of a data point from the mean of the distribution, expressed as ${(X - \mu)}/{\sigma}$. In this context, $X$ represents the contribution being evaluated, while $\mu$ and $\sigma$ denote the mean and standard deviation of the contribution distribution, respectively. FTP can be framed as a single-sample hypothesis testing task:
\begin{equation}
\mathcal{H}_{0}: \frac{X - \mu}{\sigma} \leq \lambda, \quad \text{vs.} \quad \mathcal{H}_{1}: \frac{X - \mu}{\sigma} > \lambda,
\end{equation}
where the alternative hypothesis $\mathcal{H}_{1}$ implies that the parameter behaves overconfidently, and $\lambda$ is the threshold, with $\lambda > 0$. In practice, we define $\*M^{\text{FTP}}(\*x)$ as the mask matrix customized for $\*x$ to perform FTP. The $(i,j)$-th entry of the mask matrix $\*M^{\text{FTP}}_{ij}(\*x) \in \*M^{\text{FTP}}(\*x)$ is defined as follows:
\begin{equation}
\*M_{ij}^{\text{FTP}}(\*x) = 
\begin{cases} 
1, & \text{if}\quad \dfrac{c_{j}(\*x; \*W_{ij}) - \mu_{ij}}{\sigma_{ij}} \leq \lambda, \\
0, & \text{if}\quad \dfrac{c_{j}(\*x; \*W_{ij}) - \mu_{ij}}{\sigma_{ij}} > \lambda.
\end{cases}
\end{equation}
The model output after applying FTP can be expressed as:
\begin{equation}
f^{\text{FTP}}(\*x) = \left( {\*W \odot \*M^{\text{FTP}}(\*x)} \right)^{\top} h(\*x) + \*b.
\end{equation}
Through FTP, we can adaptively prevent the abnormal increase in ID confidence caused by anomalously high contributions from overconfident parameters. This effectively reduces the risk of the model making confident yet erroneous predictions, thereby making ID data and OOD data more distinguishable.

\noindent\textbf{Overall Methods.} Both CRP and FTP pruning strategies are designed to remove parameters that negatively impact OOD detection. CRP utilizes a fixed, coarse-grained pruning pattern across all test samples to reduce interference from noisy signals. FTP applies a customized, fine-grained pruning pattern to overconfident parameters for each test sample, mitigating the risk of overconfident predictions. ITP achieves both ways to improve OOD detection. As a result, the model outputs using ITP can be expressed as follows:
\begin{equation}
\label{eq: COPP}
f^{\text{ITP}}(\*x) = \left( {\*W \odot \*M^{\text{CRP}} \odot \*M^{\text{FTP}}(\*x)} \right)^{\top} h(\*x) + \*b.
\end{equation}
Similar to previous works \cite{DBLP:conf/cvpr/AhnPK23}, we can always use the original FC layer for prediction to preserve ID accuracy with negligible additional overhead.

\begin{figure}[t]
    \centering
    \setlength{\belowcaptionskip}{-2.5pt}
    \subcaptionbox{ITP on iNaturalist benchmark}[1.0\linewidth]{
        \centering
        \begin{subfigure}{0.325\linewidth}
            \centering
            \includegraphics[width=\linewidth]{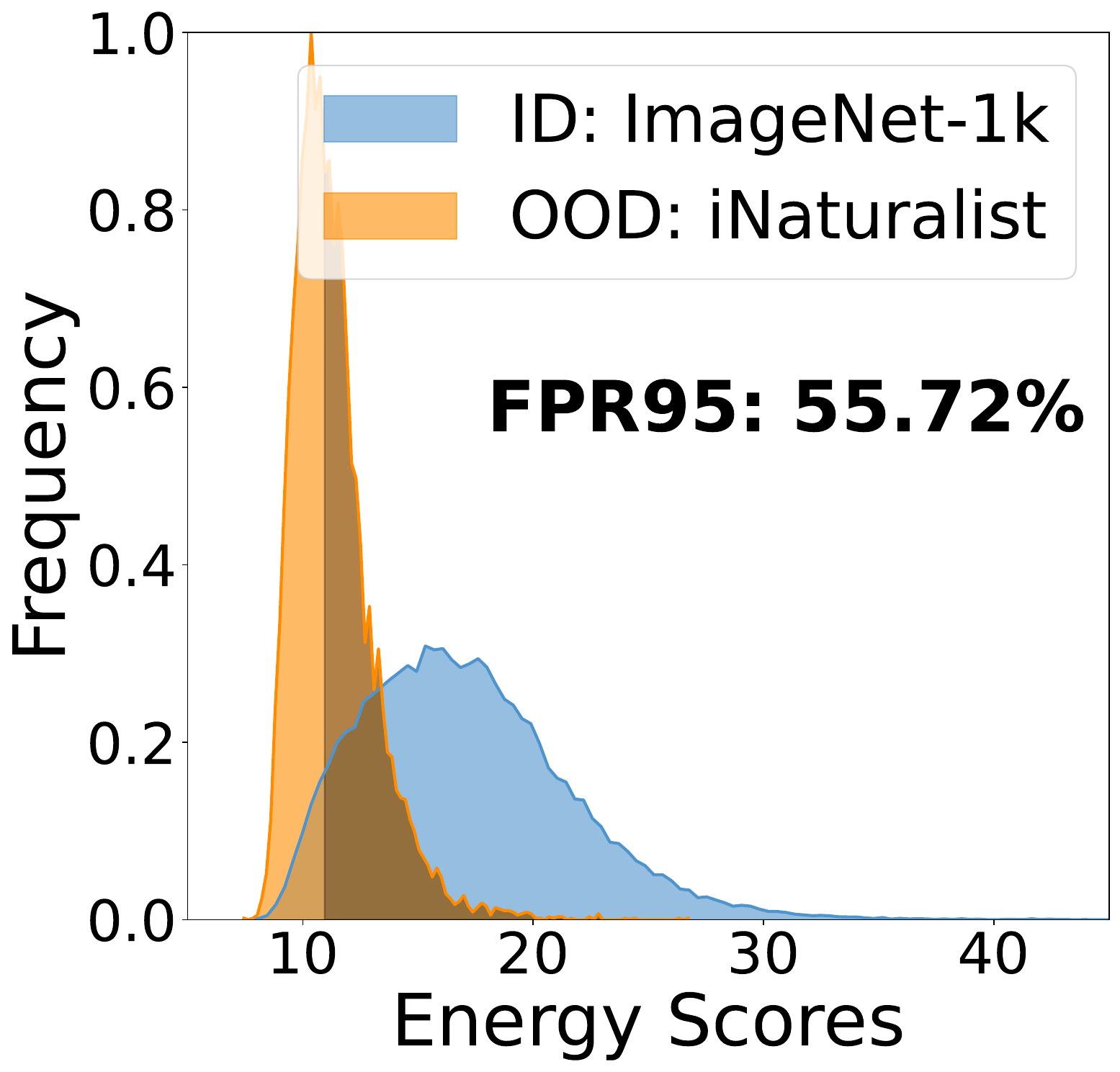}
            \caption*{Energy}
        \end{subfigure}
        \begin{subfigure}{0.325\linewidth}
            \centering
            \includegraphics[width=\linewidth]{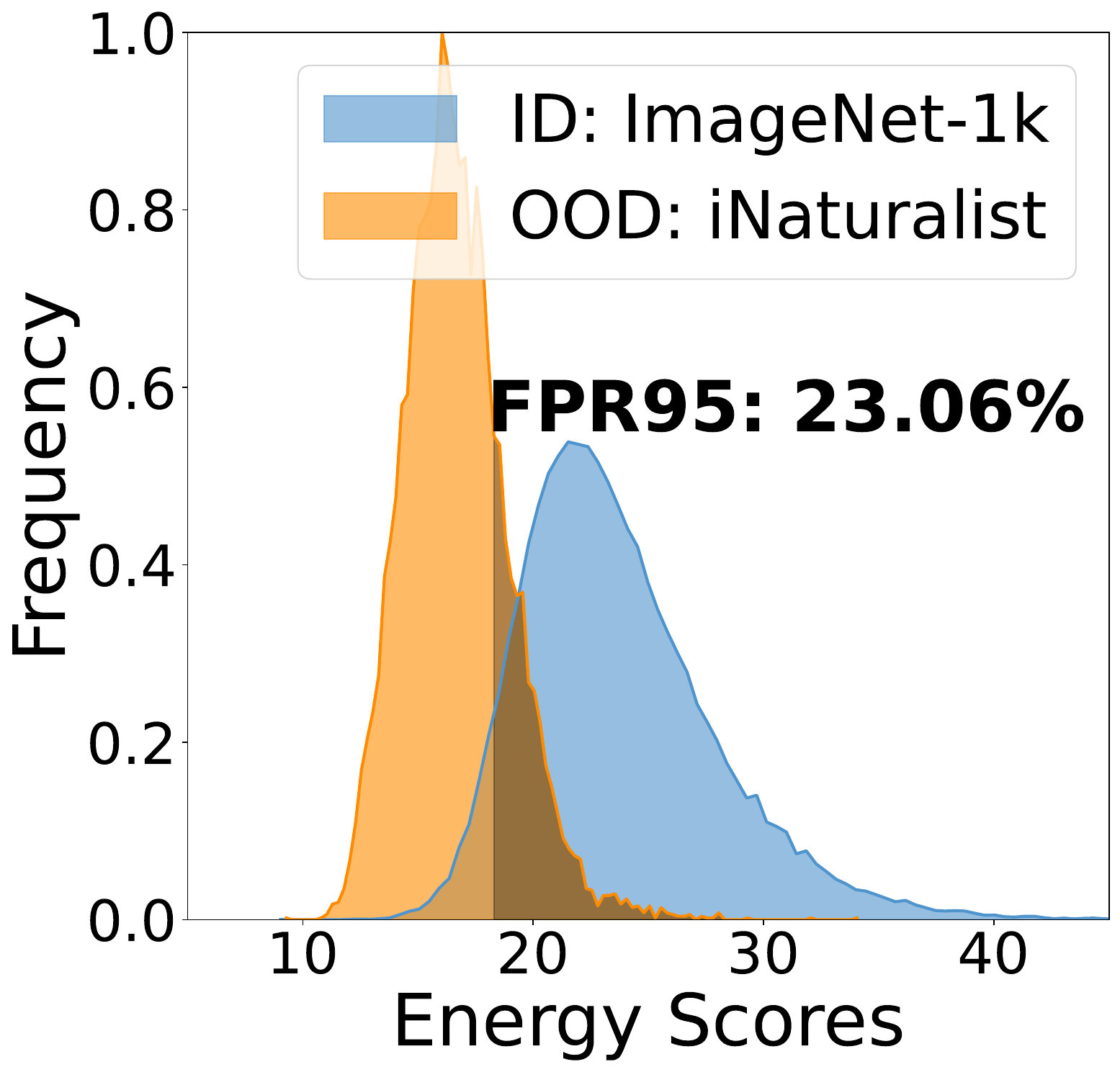}
            \caption*{CRP}
        \end{subfigure}
        \begin{subfigure}{0.325\linewidth}
            \centering
            \includegraphics[width=\linewidth]{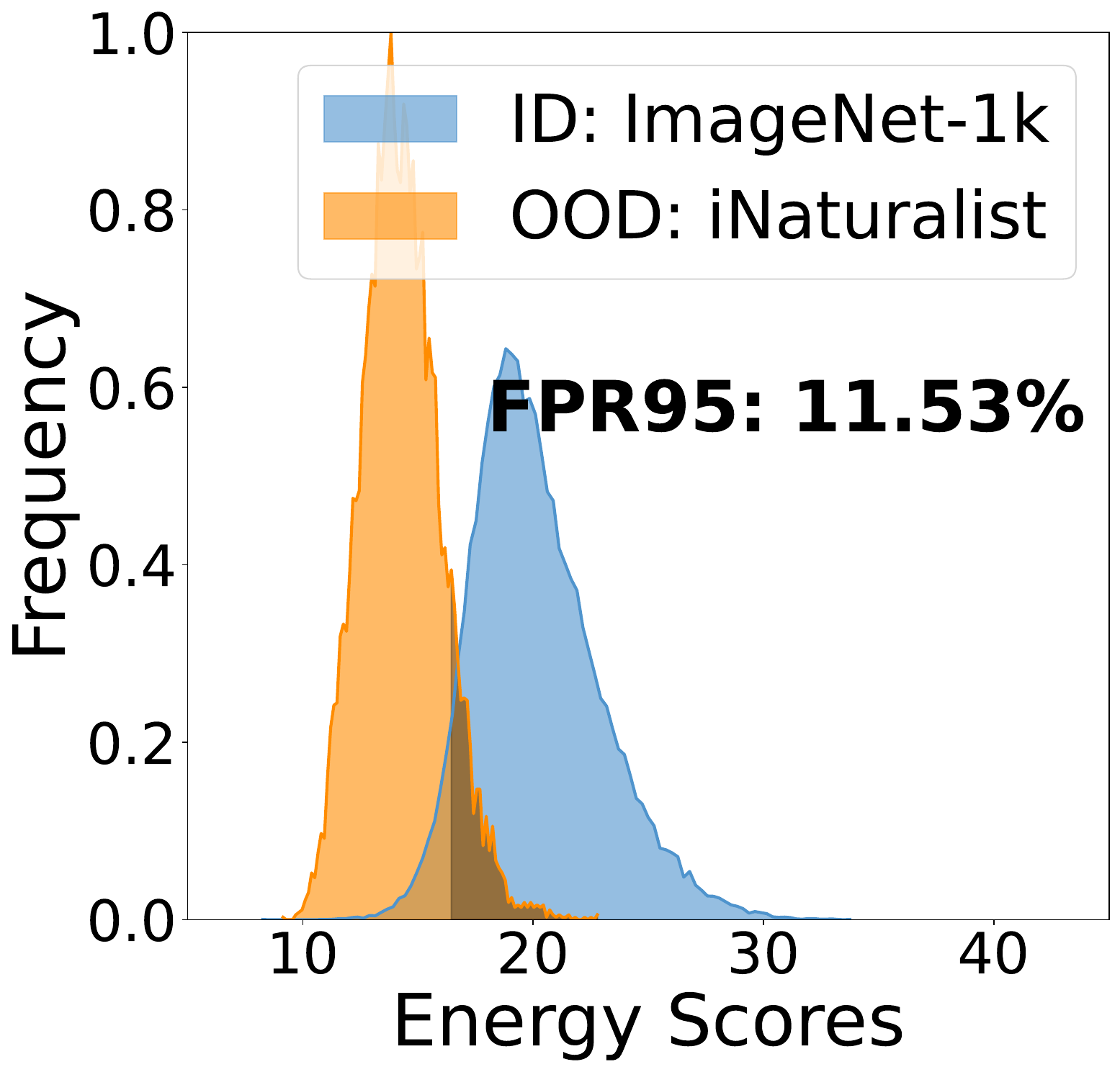}
            \caption*{ITP (CRP + FTP)}
        \end{subfigure}
    }
    \subcaptionbox{ITP on Textures benchmark}[1.0\linewidth]{
        \centering
        \begin{subfigure}{0.325\linewidth}
            \centering
            \includegraphics[width=\linewidth]{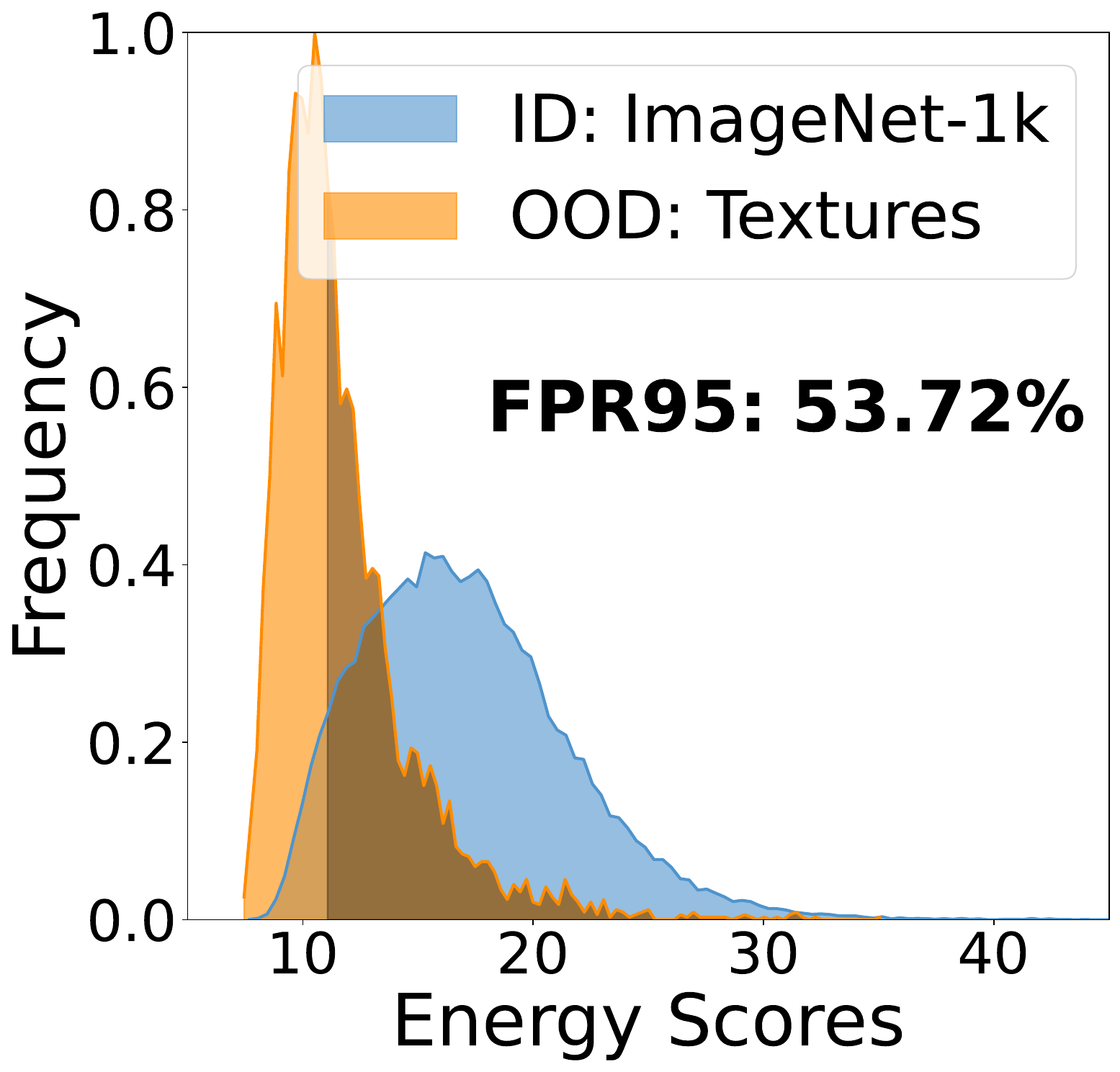}
            \caption*{Energy}
        \end{subfigure}
        \begin{subfigure}{0.325\linewidth}
            \centering
            \includegraphics[width=\linewidth]{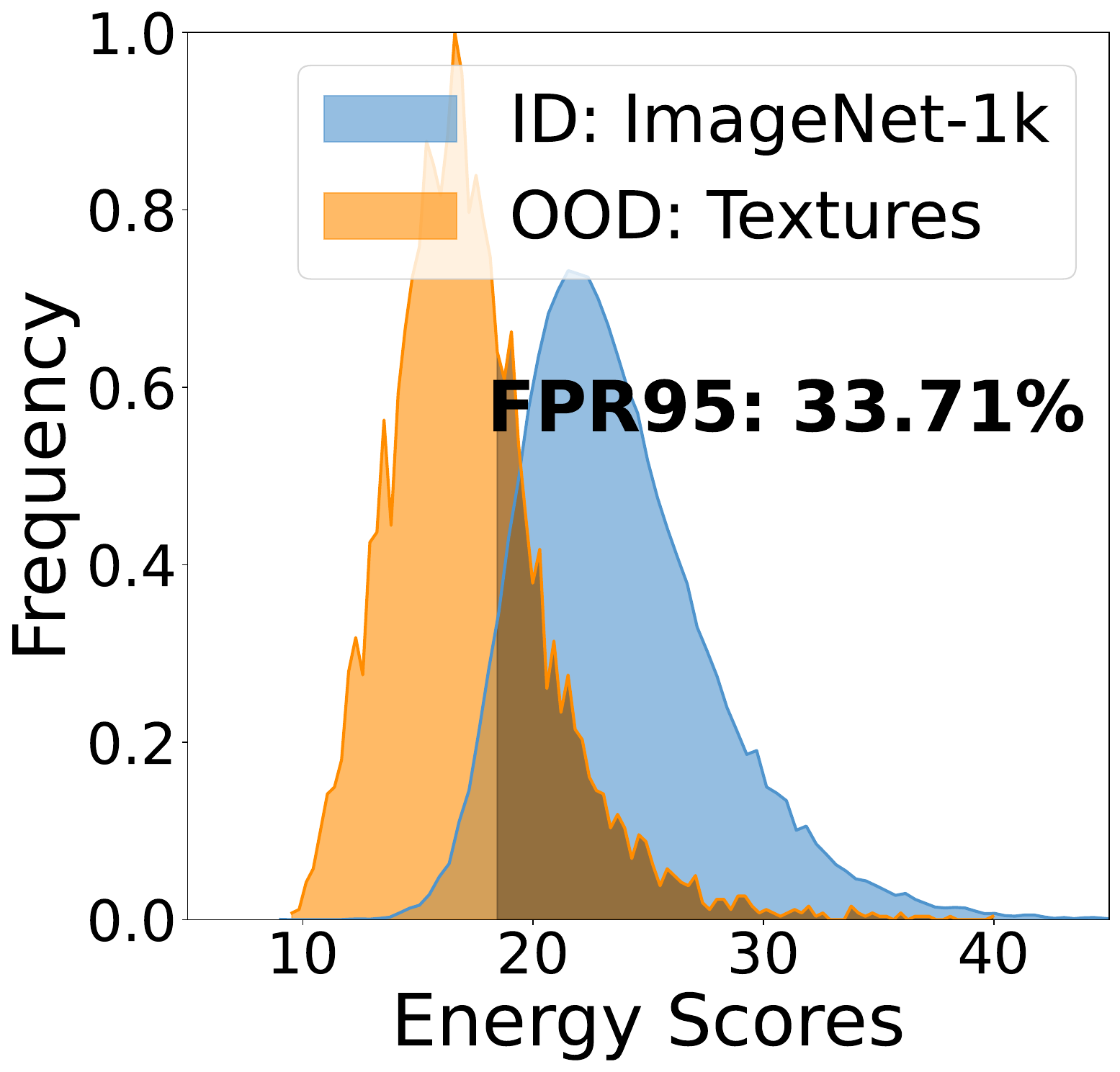}
            \caption*{CRP}
        \end{subfigure}
        \begin{subfigure}{0.325\linewidth}
            \centering
            \includegraphics[width=\linewidth]{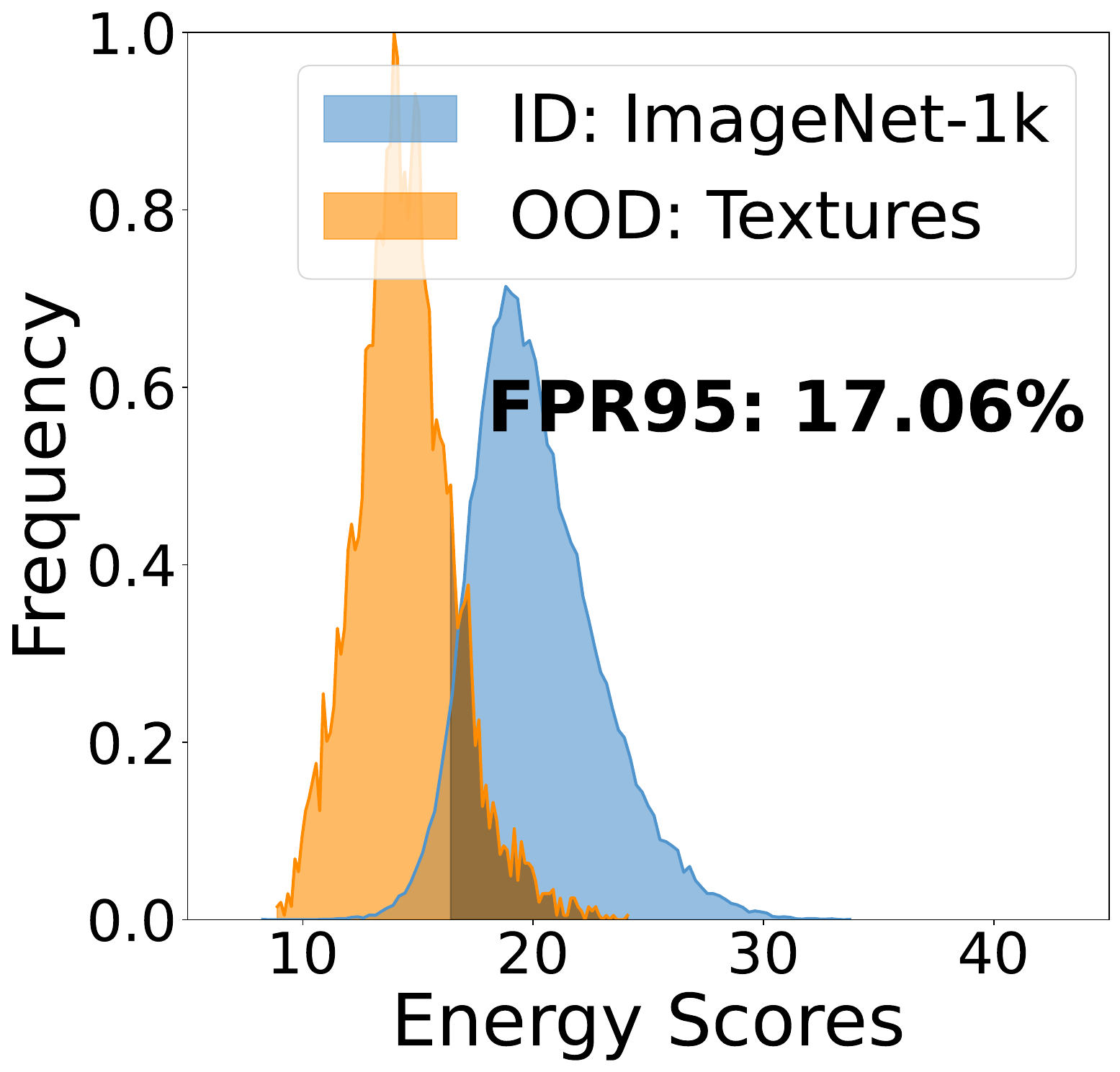}
            \caption*{ITP (CRP + FTP)}
        \end{subfigure}
    }
    \caption{Changes in OOD score distribution using ITP on (a) iNaturalist benchmark and (b) Textures benchmark.}
    \label{fig: score}
\end{figure}

\subsection{Insight Justification}
The following remarks are provided to further explain how ITP widens the gap between ID and OOD data.

\noindent\textbf{Remark 1. CRP enhances the disparity between the left tail of the OOD score distributions for ID and OOD data.}
After employing CRP to prune redundant parameters, the logits reduction for the $k$-th class is given by:
\begin{equation}
\Delta f_{j}(\*x) = \sum_{d=1}^{D}{(1-\*M^{\text{CRP}}_{dj}) \cdot c_{j}(\*x; \*W_{dj})}.
\end{equation}
CRP eliminates parameters with the least average contribution to the ID distribution. Hence, redundant parameters generally have higher contributions (manifesting as noise) to ID prediction for OOD data compared to ID data, \emph{i.e.}, 
\begin{equation}
\sum_{\*M^{\text{CRP}}_{dj} = 0}c_{j}(\*x^{\text{OOD}}; \*W_{dj}) > \sum_{\*M^{\text{CRP}}_{dj} = 0}c_{j}(\*x^{\text{ID}}; \*W_{dj}).
\end{equation}
Therefore, the reduction in logits for OOD data is greater than that for ID data $\Delta f_{j}(\*x^{\text{OOD}}) > \Delta f_{j}(\*x^{\text{ID}})$. This improves the differentiation between the left tail of the energy score distributions for ID and OOD data, due to the positive correlation between energy scores and logits. This effect is empirically validated in Figure \ref{fig: score}.

\noindent\textbf{Remark 2. FTP alleviates the overconfidence of OOD data at the right tail of the OOD score distribution.}
FTP removes the abnormally high contributions caused by overconfident parameters, thereby increasing the left skewness in the parameter contribution distribution, \emph{i.e.}, 
\begin{equation}
\begin{aligned}
&\mathbb{E}_{\*x}\left[\left(\frac{c_{j}(\*x; \*W_{ij}) - \mu_{ij}}{\sigma_{ij}}\right)^3 \cdot \*M^{\text{FTP}}_{ij}(\*x)\right] \\
&\qquad\qquad\qquad < \mathbb{E}_{\*x}\left[\left(\frac{c_{j}(\*x; \*W_{ij}) - \mu_{ij}}{\sigma_{ij}}\right)^3\right].
\end{aligned}
\end{equation}
Since parameter contributions directly determine the energy score, the left skewness of the energy score distribution will also increase accordingly. This helps reduce the risk of parameters being overly confident when handling OOD data. As a result, the overlap between the right tail of the OOD energy score distribution and the ID energy score distribution is diminished, as illustrated in Figure \ref{fig: score}.

\section{Experiments}
In this section, we first describe our experimental setup, then present the main results on multiple OOD detection benchmarks, followed by ablation studies and further analysis.

\begin{table}[t]
    \renewcommand{\arraystretch}{1.05}
    \caption{OOD detection performance on CIFAR benchmarks with DenseNet-101 as the backbone. Weight sparsification methods with ReAct (considering the count of activated features) are grouped separately for a fair comparison. All values in the table are averaged over six OOD test datasets and are percentages. The best results are in bold. $\uparrow$ indicates that larger values are better, while $\downarrow$ indicates that smaller values are better. Detailed results for each OOD dataset are provided in the Appendix.}
    \label{tab: cifar}
    \begin{tabular}{@{\hspace{0.175cm}}c@{\hspace{0.175cm}}c@{\hspace{0.175cm}}c@{\hspace{0.175cm}}c@{\hspace{0.175cm}}c@{\hspace{0.175cm}}}
    \Xhline{1.0pt}
    \multirow{3}{*}{\textbf{Method}} & \multicolumn{2}{c}{\textbf{CIFAR-10}} & \multicolumn{2}{c}{\textbf{CIFAR-100}} \\
     &  FPR95 & AUROC & FPR95 & AUROC \\
     &  $\downarrow$ & $\uparrow$ & $\downarrow$ & $\uparrow$ \\
    \Xhline{0.5pt}
    MSP & 48.73 & 92.46 & 80.13 & 74.36 \\
    Energy & 26.55 & 94.57 & 68.45 & 81.19 \\
    ODIN & 24.57 & 93.71 & 58.14 & 84.49 \\
    ReAct & 26.45 & 94.67 & 62.27 & 84.47 \\
    DICE & 20.83 & 95.24 & 49.72 & 87.23 \\
    OPNP & 22.07 & 95.14 & 51.79 & 87.20  \\
    LAPS & 19.40 & 96.10 & 50.50 & 88.07 \\
    \textbf{ITP (Ours)} & \textbf{16.72} & \textbf{96.64} & \textbf{35.03} & \textbf{91.39}\\
    \Xhline{0.1pt}
    DICE + ReAct & 16.48 & 96.64 & 49.57 & 85.07\\
    OPNP + ReAct & 18.46 & 96.35 & 42.98 & 88.55  \\
    LINe (w/ ReAct) & 14.72 & 96.99 & 35.67 & 88.67 \\
    \textbf{ITP + ReAct (Ours)} & \textbf{14.50} & \textbf{97.13} & \textbf{30.13} & \textbf{91.91} \\
    \Xhline{1.0pt}
    \end{tabular}
\end{table}

\begin{table*}[t]
    \renewcommand{\arraystretch}{1.05}
    \caption{OOD detection performance on ImageNet with ResNet-50 as the backbone.}
    \centering
    \label{tab: imagenet}
    \begin{tabular}{@{\hspace{0.30cm}}c@{\hspace{0.30cm}}c@{\hspace{0.30cm}}c@{\hspace{0.30cm}}c@{\hspace{0.30cm}}c@{\hspace{0.30cm}}c@{\hspace{0.30cm}}c@{\hspace{0.30cm}}c@{\hspace{0.30cm}}c@{\hspace{0.30cm}}c@{\hspace{0.30cm}}c@{\hspace{0.30cm}}}
    \Xhline{1.0pt}
     \multirow{4}{*}{\textbf{Method}} & \multicolumn{8}{c}{\textbf{OOD Datasets}} & \multicolumn{2}{c}{\multirow{2}{*}{\textbf{Average}}} \\
    \cline{2-9}
    & \multicolumn{2}{c}{\textbf{iNaturalist}} & \multicolumn{2}{c}{\textbf{SUN}} & \multicolumn{2}{c}{\textbf{Places}} & \multicolumn{2}{c}{\textbf{Textures}} \\
    & FPR95 & AUROC & FPR95 & AUROC & FPR95 & AUROC & FPR95 & AUROC & FPR95 & AUROC \\
    & $\downarrow$ & $\uparrow$& $\downarrow$ & $\uparrow$& $\downarrow$ & $\uparrow$& $\downarrow$ & $\uparrow$& $\downarrow$ & $\uparrow$ \\
    \Xhline{0.5pt}
    MSP & 54.99 & 87.74 & 70.83 & 80.86 & 73.99 & 79.76 & 68.00 & 79.61 & 66.95 & 81.99 \\
    Energy  & 55.72 & 89.95 & 59.26 & 85.89 & 64.92 & 82.86 & 53.72 & 85.99 & 58.41 & 86.17 \\
    ODIN & 47.66 & 89.66 & 60.15 & 84.59 & 67.89 & 81.78 & 50.23 & 85.62 & 56.48 & 85.41 \\
    ReAct & 20.38 & 96.22 & 24.20 & 94.20 & 33.85 & 91.58 & 47.30 & 89.80 & 31.43 & 92.95 \\ 
    DICE & 25.63 & 94.49 & 35.15 & 90.83 & 46.49 & 87.48 & 31.72 & 90.30 & 34.75 & 90.77 \\
    OPNP & 18.89 & 96.03 & 18.50 & 95.62 & 30.14 & 93.46 & 36.17 & 91.70 & 25.93 & 94.20 \\
    LAPS & 12.72 & 97.50 & \textbf{15.81} & \textbf{96.18} & \textbf{24.71} & \textbf{93.64} & 41.49 & 91.81 & 23.68 & \textbf{94.78} \\
    \textbf{ITP (Ours)} & \textbf{11.53} & \textbf{97.83} & 25.82 & 93.58 & 35.63 & 90.75 & \textbf{17.06} & \textbf{96.03} & \textbf{22.51} & 94.55 \\
    \Xhline{0.5pt}
    DICE + ReAct & 18.64 & 96.24 & 25.45 & 93.94 & 36.86 & 90.67 & 28.07 & 92.74 & 27.25 & 93.40\\
    OPNP + ReAct & 14.72 & 96.78 & 19.73 & \textbf{95.65} & 30.23 & \textbf{93.34} & 27.78 & 94.13 & 23.12 & 94.98\\
    LINe (w/ ReAct) & 12.26 & 97.56 & \textbf{19.48} & 95.26 & \textbf{28.52} & 92.85 & 22.54 & 94.44 & 20.70 & 95.03 \\
    \textbf{ITP + ReAct (Ours)}  & \textbf{9.78} & \textbf{98.02} & 22.82 & 94.47 & 30.87 & 92.03 & \textbf{18.09} & \textbf{95.98} & \textbf{20.39} & \textbf{95.13} \\
    \Xhline{1.0pt}
    \end{tabular}
\end{table*}

\subsection{Experimental Setup}
In line with other OOD literature \cite{DBLP:conf/eccv/SunL22a}, we evaluate our methods both on the small-scale CIFAR benchmarks and the large-scale ImageNet benchmark\footnote{Code is available at \url{https://github.com/njustkmg/AAAI25-ITP}}. Moreover, we provide a further evaluation of our proposal on the OpenOOD v1.5 benchmark \cite{DBLP:journals/corr/abs-2306-09301} in the appendix. We default to using the entire training set for estimating the parameter contribution distribution.

\noindent\textbf{CIFAR.} 
We use CIFAR-10 and CIFAR-100 \cite{Krizhevsky_2009} as ID datasets and consider six OOD datasets: SVHN \cite{Netzer_Wang_Coates_Bissacco_Wu_Ng_2011}, Textures \cite{DBLP:conf/cvpr/CimpoiMKMV14}, iSUN \cite{xu2015turkergaze}, LSUN-Resize \cite{DBLP:journals/corr/YuZSSX15}, LSUN-Crop \cite{DBLP:journals/corr/YuZSSX15}, and Places365 \cite{DBLP:journals/pami/ZhouLKO018}. For consistency with previous work \cite{DBLP:conf/eccv/SunL22a}, we use the same model architecture and pre-trained weights, namely DenseNet-101 \cite{DBLP:conf/cvpr/HuangLMW17}.

\noindent\textbf{ImageNet.} 
For the large-scale ImageNet experiments, we use the ImageNet-1k as the ID dataset and consider (subsets of) iNaturalist \cite{DBLP:conf/cvpr/HornASCSSAPB18}, Places \cite{DBLP:journals/pami/ZhouLKO018}, SUN \cite{DBLP:conf/cvpr/XiaoHEOT10}, and Textures \cite{DBLP:conf/cvpr/CimpoiMKMV14} with non-overlapping categories from ImageNet-1k as OOD datasets. We adopt the widely used ResNet-50 \cite{DBLP:conf/cvpr/HeZRS16} model architectures, and we obtain the pre-trained weights from the torchvision library.

\noindent\textbf{Baselines.} 
We compare ITP with the most competitive OOD detection methods: MSP \cite{DBLP:conf/iclr/HendrycksG17}, Energy \cite{DBLP:conf/nips/LiuWOL20}, ODIN \cite{DBLP:conf/iclr/LiangLS18}, ReAct \cite{DBLP:conf/nips/SunGL21}, DICE \cite{DBLP:conf/eccv/SunL22a}, LINe \cite{DBLP:conf/cvpr/AhnPK23}, OPNP \cite{DBLP:conf/nips/ChenFLCTY23}, and LAPS \cite{DBLP:conf/aaai/HeYHWSYLG24}. Moreover, to align with standard sparsification practices, we also report the results of a comparison with ReAct (e.g., ITP + ReAct). In particular, LINe integrates ReAct within its framework, which we refer to as ``LINE (w/ ReAct)'' in the table. All methods are post-hoc and can be directly applied to pre-trained models.

\noindent\textbf{Evaluation Metric.} 
We adopt two threshold-free metrics for evaluation. FPR95: the false positive rate of OOD data at 95\% true positive rate of ID data. AUROC: the area under the receiver operating characteristic curve.

\begin{table}[t]
    \renewcommand{\arraystretch}{1.05}
    \caption{Ablation study for our proposed method. All values are percentages and averaged over multiple OOD datasets.}
    \label{tab: ablation}
    \centering
    \begin{tabular}{@{\hspace{0.25cm}}c@{\hspace{0.25cm}}c@{\hspace{0.25cm}}c@{\hspace{0.25cm}}c@{\hspace{0.25cm}}c@{\hspace{0.25cm}}}
    \Xhline{1.0pt}
    \textbf{Dataset} &\textbf{CRP} &\textbf{FTP} & \textbf{FPR95} & \textbf{AUROC} \\ 
    & & & $\downarrow$ & $\uparrow$\\
    \Xhline{0.5pt}
    \multirow{4}{*}{\makecell[c]{CIFAR-10\\ DenseNet-101}} & $\times$ & $\times$ & 26.55 & 94.57\\ 
    & $\checkmark$ & $\times$  & 21.29 & 95.09 \\
    & $\times$ & $\checkmark$  & 19.16 & 96.36 \\
    & $\checkmark$ & $\checkmark$  & \textbf{16.96} & \textbf{96.59} \\
    \Xhline{0.5pt}
    \multirow{4}{*}{\makecell[c]{CIFAR-100\\ DenseNet-101}} & $\times$ & $\times$  & 68.45 & 81.19\\ 
    & $\checkmark$ & $\times$  & 53.60 & 85.33 \\
    & $\times$ & $\checkmark$  & 53.73 & 87.56 \\
    & $\checkmark$ & $\checkmark$  & \textbf{35.03} & \textbf{91.39} \\
    \Xhline{0.5pt}
    \multirow{4}{*}{\makecell[c]{ImageNet-1k\\ResNet-50}} & $\times$ & $\times$  & 58.41 & 86.17\\ 
    & $\checkmark$ & $\times$  & 33.96 & 91.26 \\
    & $\times$ & $\checkmark$  & 51.38 & 88.68 \\
    & $\checkmark$ & $\checkmark$  & \textbf{22.51} & \textbf{94.55} \\
    \Xhline{1.0pt}
    \end{tabular}
\end{table}

\subsection{Main Results}
In this section, we report the performance of our ITP on commonly used CIFAR benchmarks and the more realistic and challenging ImageNet benchmark. Baseline results are sourced from \cite{DBLP:conf/cvpr/AhnPK23,DBLP:conf/nips/ChenFLCTY23,DBLP:conf/aaai/HeYHWSYLG24}, with additional baselines (\emph{e.g.}, LAPS, OPNP, and OPNP + ReAct on CIFAR) reproduced by us. 

For the CIFAR benchmarks, Table \ref{tab: cifar} lays out the performance of OOD detection on the CIFAR-10 and CIFAR-100, respectively. As we can see, the proposed ITP outperforms all baselines considered and achieves state-of-the-art performance. In CIFAR-100, ITP reduces FPR95 by 33.42\% compared to the energy baseline, showing the effectiveness of our proposal with the same OOD scoring function. Remarkably, ITP + ReAct outperforms the most competitive method LINe by 5.54\% in FPR95 and 3.24\% in AUROC, highlighting the importance of further fine-grained pruning of overconfident parameters.

For the large-scale ImageNet benchmark, Table \ref{tab: imagenet} reports detailed performances for each OOD dataset and the average over the four datasets. Our proposed method, ITP, outperforms recent approaches such as DICE, OPNP, and LAPS, achieving the best performance among the baseline methods with an FPR95 of 22.51\%. Moreover, the combination of ITP and ReAct outperforms recent approaches DICE + ReAct, OPNP + ReAct, and LINe. The experimental results demonstrate that our ITP is state-of-the-art and effective for OOD detection on large-scale real-world datasets.

\begin{table}[t]
    \caption{Impact of varying hyperparameters on FPR95. We use ImageNet-1k as the ID dataset and ResNet-50 as the pre-trained model. All values are percentages and averaged over four OOD datasets.}
    \label{tab: hyperparameter}
    \centering
    \renewcommand{\arraystretch}{1.05}
    \begin{tabular}{@{\hspace{0.20cm}}c@{\hspace{0.20cm}}c@{\hspace{0.20cm}}c@{\hspace{0.20cm}}c@{\hspace{0.20cm}}c@{\hspace{0.20cm}}c@{\hspace{0.20cm}}}
    \Xhline{1.0pt}
    & $p = 10$ & $p = 30$ & $p = 50$ & $p = 70$ & $p = 90$\\
    \Xhline{0.5pt}
    $\lambda = 0.5$ & 73.19 & 34.19 & 33.70 & 33.75 & 35.96 \\
    $\lambda = 1.0$ & 33.18 & 25.44 & 26.83 & 27.26 & 30.96 \\
    $\lambda = 1.5$ & 26.30 & \textbf{22.51} & 24.18 & 24.67 & 29.50\\
    $\lambda = 2.0$ & 27.58 & 24.69 & 25.95 & 26.45 & 31.59 \\
    $\lambda = 3.0$ & 31.95 & 29.14 & 30.37 & 30.73 & 36.53 \\
    $\lambda = 5.0$ & 35.90 & 33.00 & 34.14 & 34.52 & 40.23 \\
    \Xhline{1.0pt}
    \end{tabular}
\end{table}

\subsection{Ablation Study}

\noindent\textbf{Ablation on proposed pruning strategies.} 
To fully demonstrate the impact of different granularity pruning strategies in ITP, we conduct a comprehensive empirical analysis on CIFAR-10, CIFAR-100, and ImageNet-1k, and report the results in Table \ref{tab: ablation}. As shown in the table, both FTP and CRP improve performance, and ITP further markedly boosts OOD detection performance by integrating these two coarse-grained and fine-grained pruning strategies. However, the improvement of FTP on ResNet-50 is less pronounced, likely due to its larger feature space (2048 dimensions) compared to DenseNet-101 (342 dimensions). This larger space allows noise to dominate and interfere with FTP. The significant improvement observed when FTP is applied after noise removal with CRP supports this explanation. The ablation study verifies the effectiveness of the two strategies and demonstrates that they mutually enhance and complement each other.

\noindent\textbf{Effect of the Hyperparameter.} 
Table \ref{tab: hyperparameter} shows the results of varying the percentile $p$ used for pruning redundant parameters and the threshold $\lambda$ for identifying overconfident parameters. The optimal performance is observed at $p = 30$ and $\lambda = 1.5$, achieving an FPR95 of 22.51\%. Conversely, we notice that selecting excessively large values for $p$ and overly small values for $\lambda$ can lead to the erroneous removal of critical parameters, thereby impairing OOD detection.

\begin{figure}[t]
\centering
\includegraphics[width=0.9\linewidth]{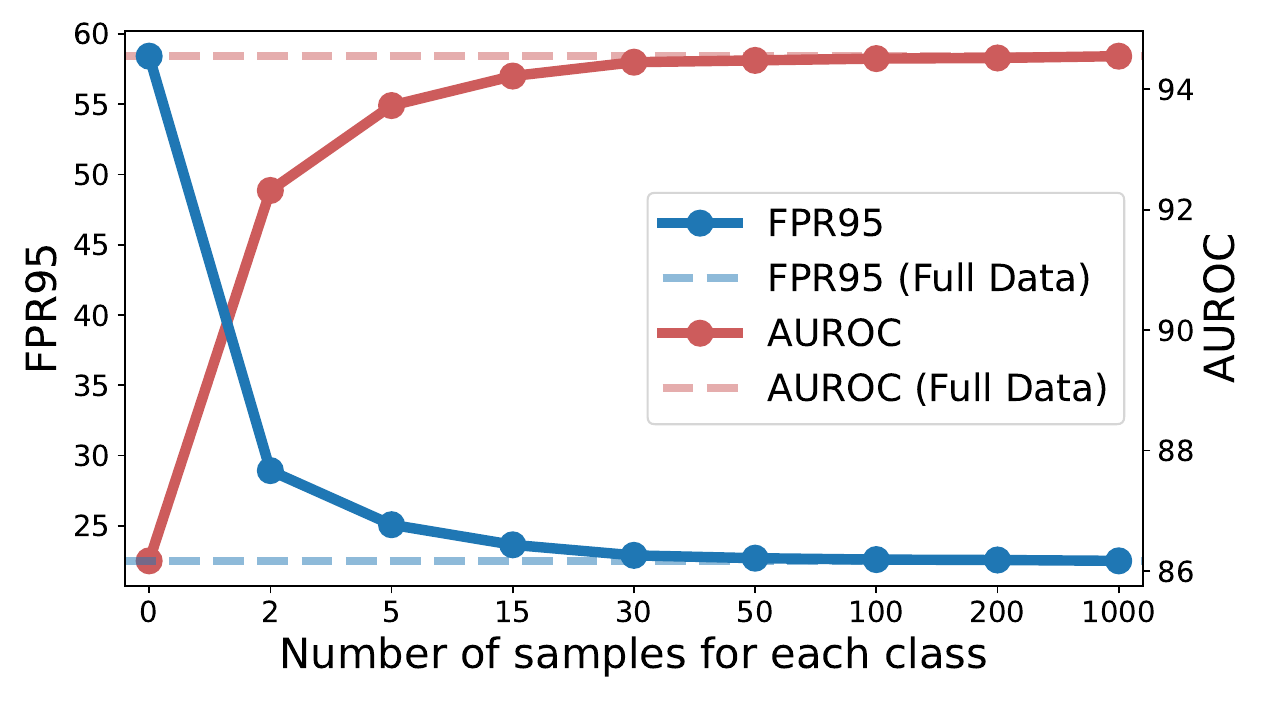}
\caption{The FFPR95 and AUROC with different number of training samples on ImageNet benchmark. The results are averaged over five independent runs.}
\label{fig: amount}
\end{figure}

\noindent\textbf{Effect of the Amount of Training Samples.} 
In Figure \ref{fig: amount}, we show the effect of utilizing different numbers of training samples to estimate the parameter contribution distribution. Remarkably, even with just two samples per class, ITP can significantly improve OOD detection performance, resulting in a drastic 29.48\% reduction in FPR95 compared to the energy baseline (without pruning). Furthermore, empirical evidence suggests that using only 30 samples per class can yield performance nearly equivalent to that achieved with the full dataset. Therefore, to reduce computational overhead, it is feasible to use a suitably sized subset of the training set (\emph{e.g.}, 30 samples per class) for distribution estimation, while still achieving comparable performance.

\subsection{Further Analysis}

\begin{table}[t]
    \centering
    \renewcommand{\arraystretch}{1.05}
    \captionof{table}{Comparison of preprocessing overhead. We assess the preprocessing overhead by averaging the preprocessing times measured across three runs on ImageNet-1k. ``ITP (30)'' denotes that only 30 images per class are used for ITP while maintaining the original performance (see Figure \ref{fig: amount}).}
    \label{tab: overhead}
    \begin{tabular}{@{\hspace{0.22cm}}c@{\hspace{0.22cm}}c@{\hspace{0.22cm}}c@{\hspace{0.22cm}}c@{\hspace{0.22cm}}}
        \Xhline{1.0pt}
        \multirow{2}*{\textbf{Method}} & \multirow{2}*{\makecell[c]{\textbf{Preprocessing} \\ \textbf{Time (hours)}}} & \multirow{2}*{\makecell[c]{\textbf{Additional} \\ \textbf{Backpropagation}}} & \multirow{2}*{\makecell[c]{\textbf{Batch} \\ \textbf{Support}}} \\
        & & & \\
        \Xhline{0.5pt}
        OPNP  & 8.7940 & $\checkmark$ & $\times$ \\
        LINe  & 7.3096 & $\checkmark$ & $\times$\\
        \Xhline{0.5pt}
        ITP  & 0.2411 & $\times$ & $\checkmark$ \\
        ITP (30)  & $ \textbf{0.0073} $ & $\times$ & $\checkmark$ \\
        \Xhline{1.0pt}
    \end{tabular}
\end{table}

\noindent\textbf{Analysis of Preprocessing Overhead.}
Table \ref{tab: overhead} compares the preprocessing overhead of ITP with the most competitive weight sparsification methods: LINe and OPNP. In contrast to ITP, both LINe \cite{DBLP:conf/cvpr/AhnPK23} and OPNP \cite{DBLP:conf/nips/ChenFLCTY23} require additional backpropagation to compute gradient information and lack support for batch processing. The results indicate that our proposal demonstrates a significant advantage in terms of preprocessing overhead. Notably, with only 30 samples per class, we can further substantially reduce the overhead while maintaining comparable performance (see Figure \ref{fig: amount}). Therefore, ITP is efficient and well-suited for real-world applications.

\begin{table}[t]
    \centering
    \renewcommand{\arraystretch}{1.05}
    \captionof{table}{ITP on other OOD scores. We use ResNet-50 as the pre-trained model and ImageNet-1k as the ID dataset. The results are averaged over four OOD datasets.}
    \label{tab: other}
    \begin{tabular}{ccc}
    \Xhline{1.0pt}
    \textbf{Method} & \textbf{FPR95} & \textbf{AUROC} \\ 
    & $\downarrow$ & $\uparrow$\\
    \Xhline{0.5pt}
    MSP & 66.95 & 81.99\\
    MSP + \textbf{ITP} & \textbf{62.44} & \textbf{82.99} \\
    \Xhline{0.5pt}
    ODIN & 56.48 & 85.41\\ 
    ODIN + \textbf{ITP}  & \textbf{42.32} & \textbf{90.10}\\ 
    \Xhline{0.5pt}
    GradNorm  & 36.49 & 90.18 \\ 
    GradNorm + \textbf{ITP}  & \textbf{29.93} & \textbf{92.13}\\
    \Xhline{0.5pt}
    MLS & 58.05 & 87.00 \\ 
    MLS + \textbf{ITP} & \textbf{26.43} & \textbf{93.45}\\ 
    \Xhline{0.5pt}
    Energy & 58.41 & 86.17\\ 
    Energy + \textbf{ITP} & \textbf{22.51} & \textbf{94.55}\\ 
    \Xhline{1.0pt}
    \end{tabular}
\end{table}

\noindent\textbf{Compatibility with Other OOD Scores.}
Table \ref{tab: other} presents the OOD detection performance of ITP using various OOD scoring methods, including MSP \cite{DBLP:conf/iclr/HendrycksG17}, ODIN \cite{DBLP:conf/iclr/LiangLS18}, GradNorm \cite{DBLP:conf/nips/HuangGL21}, MLS \cite{DBLP:conf/icml/HendrycksBMZKMS22}, and Energy \cite{DBLP:conf/nips/LiuWOL20}. Our ITP consistently improves FPR95 and AUROC across different OOD scores. In particular, ITP can effectively complement gradient-based methods such as GradNorm. These results indicate that the parameters our ITP selectively used are also applicable to other OOD scores and show strong compatibility.

\section{Conclusion}
In this paper, we reveal that parameters important for ID data prediction are not always beneficial for OOD detection. To address this issue, we propose a parameter pruning method called ITP, which utilizes class-specific parameter contribution distributions for post-hoc OOD detection. ITP is based on two powerful pruning strategies: CRP performs coarse-grained pruning to remove redundant parameters, while FTP executes fine-grained pruning to eliminate overconfident parameters. Experimental results show that our ITP method significantly improves OOD detection performance and can be integrated with a wide range of other OOD scoring methods. We hope our work can raise more attention to the importance of test parameter pruning for OOD detection.

\section{Acknowledgments}
National Key RD Program of China (2022YFF0712100), NSFC (62276131), Natural Science Foundation of Jiangsu Province of China under Grant (BK20240081), the Fundamental Research Funds for the Central Universities (No.30922010317), Key Laboratory of Target Cognition and Application Technology (2023-CXPT-LC-005).

\bibliography{aaai25}
\clearpage

\appendix
\section{Details of Parameter Contribution}
The contribution of the model weight parameter $\mathbf{W}_{ij}$ in the last layer to the $k$-th class is defined as the difference between the model's output for the $k$-th class when $\mathbf{W}_{ij}$ is active and when $\mathbf{W}_{ij}$ is set to zero:
\begin{equation}
c_{k}(\mathbf{x}; \mathbf{W}_{ij}) = f_{k}(\mathbf{x}) - f_{k}(\mathbf{x}; \mathbf{W}_{ij} = 0).
\label{apd: contribution}
\end{equation}
Here, $f_{k}(\mathbf{x})$ can be expressed as follows according to the definition of matrix multiplication, \emph{i.e.}, 
\begin{equation}
f_{k}(\mathbf{x}) = \sum\nolimits_{d=1}^D \mathbf{W}_{dk} \cdot h_{d}(\mathbf{x}) + \mathbf{b}_{k},
\end{equation}
where $D$ is the dimension of the penultimate layer's features, $h_{d}(\mathbf{x})$ represents the activation value of the $d$-th neuron, and $\mathbf{b}_{k}$ is the bias term associated with the $k$-th output. To determine the contribution of $\mathbf{W}_{ij}$ to the $k$-th class, we consider two cases:

(1) When $k = j$, the effect of setting $\mathbf{W}_{ij}$ to zero on the output $f_{k}(\mathbf{x})$ is:
\begin{equation}
f_{k}(\mathbf{x}; \mathbf{W}_{ij} = 0) = \sum\nolimits_{d=1,d \ne i}^D \mathbf{W}_{dk} \cdot h_{d}(\mathbf{x}) + \mathbf{b}_{k},
\end{equation}
where the contribution from the weight $\mathbf{W}_{ij}$ corresponding to conditions $k = j$ and $d = i$ is removed.

(2) When $k \ne j$, the effect of setting $\mathbf{W}_{ij}$ to zero on the output $f_{k}(\mathbf{x})$ is:
\begin{equation}
f_{k}(\mathbf{x}; \mathbf{W}_{ij} = 0) = \sum\nolimits_{d=1}^D \mathbf{W}_{dk} \cdot h_{d}(\mathbf{x}) + \mathbf{b}_{k},
\end{equation}
since $\mathbf{W}_{ij}$ does not contribute to the computation of $f_{k}(\mathbf{x})$ when $k \ne j$.

Therefore, according to Equation \ref{apd: contribution}, the contribution of $\mathbf{W}_{ij}$ to the $k$-th class can be succinctly expressed as:
\begin{equation}
\label{eq: pc}
f_{k}(\mathbf{x}) - f_{k}(\mathbf{x}; \mathbf{W}_{ij} = 0) =
\begin{cases} 
\mathbf{W}_{ij} \cdot h_{i}(\mathbf{x}), & \text{if } k = j, \\
0, & \text{if } k \ne j.
\end{cases}
\end{equation}
This result shows that $\mathbf{W}_{ij}$ directly affects the $k$-th class only when $k$ matches $j$, and has no impact on other classes.

\section{Distribution of Parameter Contribution}
\begin{figure*}[t]
    \centering
    \setlength{\belowcaptionskip}{-2.5pt}
    \subcaptionbox{DenseNet-101}[1.0\linewidth]{
        \centering
        \begin{subfigure}{0.195\linewidth}
            \centering
            \includegraphics[width=\linewidth]{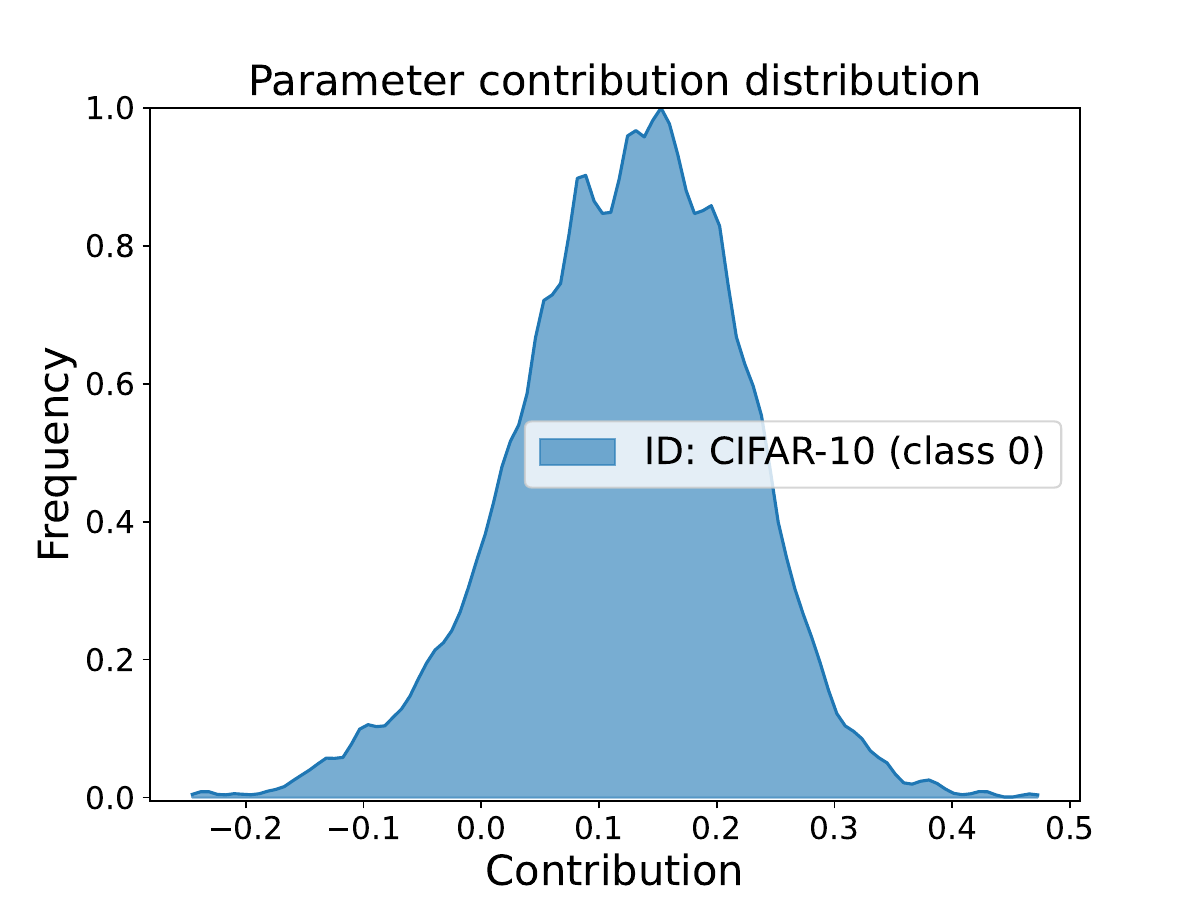}
        \end{subfigure}
        \begin{subfigure}{0.195\linewidth}
            \centering
            \includegraphics[width=\linewidth]{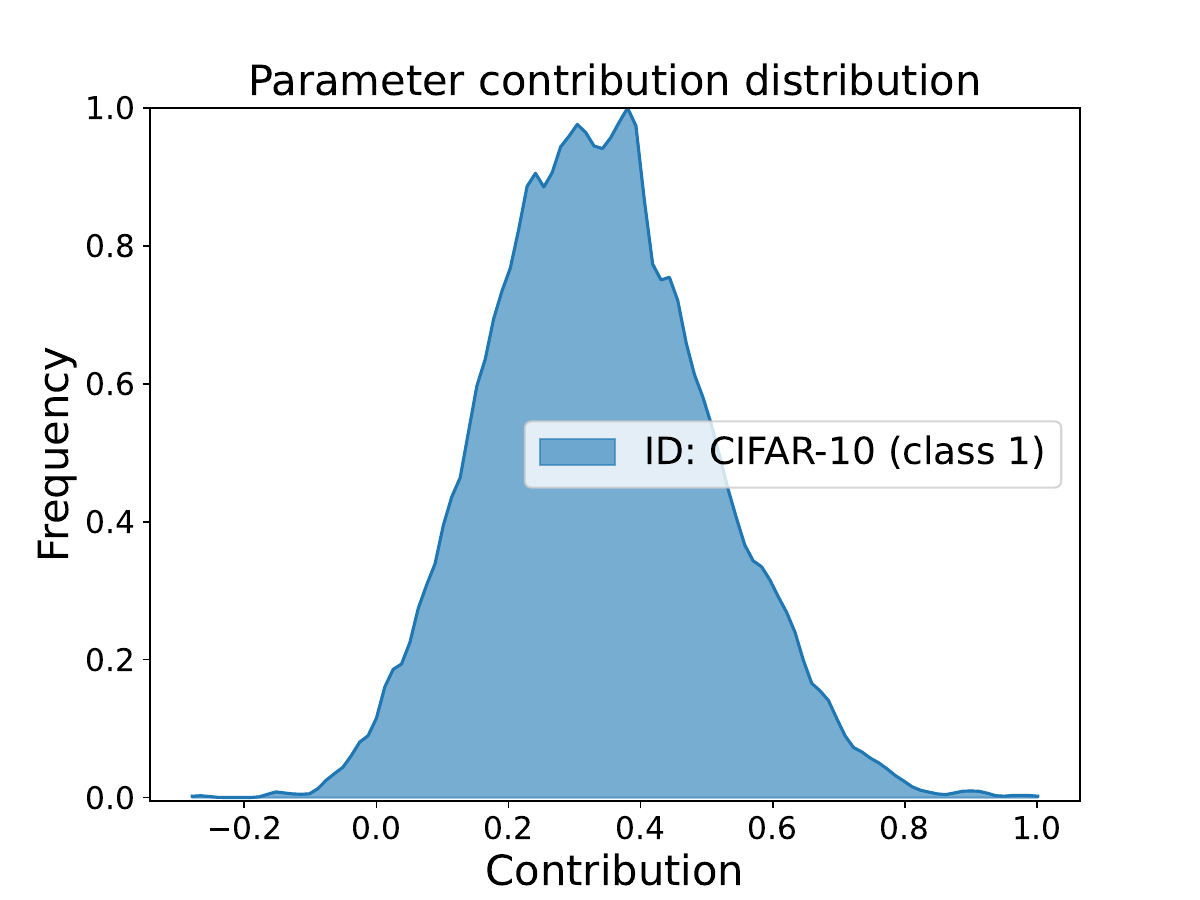}
        \end{subfigure}
        \begin{subfigure}{0.195\linewidth}
            \centering
            \includegraphics[width=\linewidth]{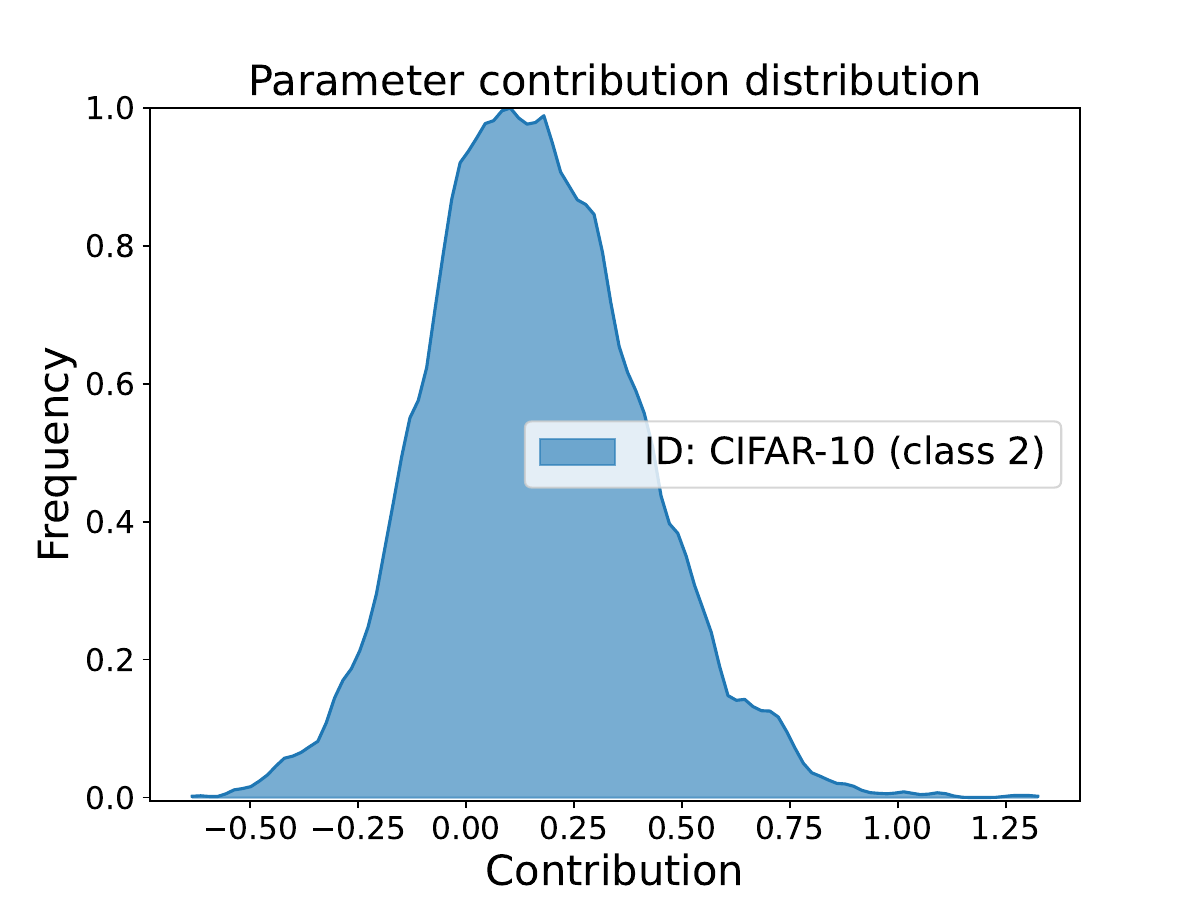}
        \end{subfigure}
        \begin{subfigure}{0.195\linewidth}
            \centering
            \includegraphics[width=\linewidth]{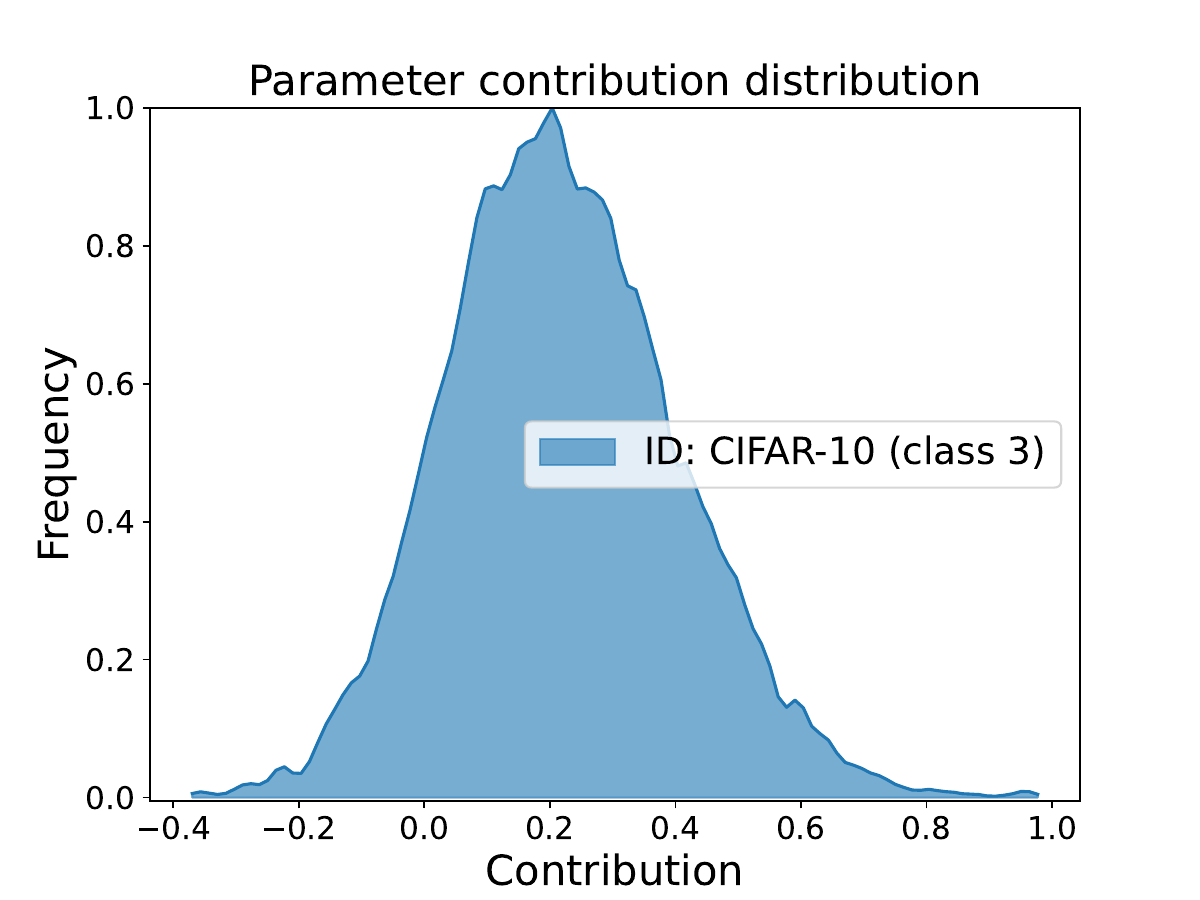}
        \end{subfigure}
        \begin{subfigure}{0.195\linewidth}
            \centering
            \includegraphics[width=\linewidth]{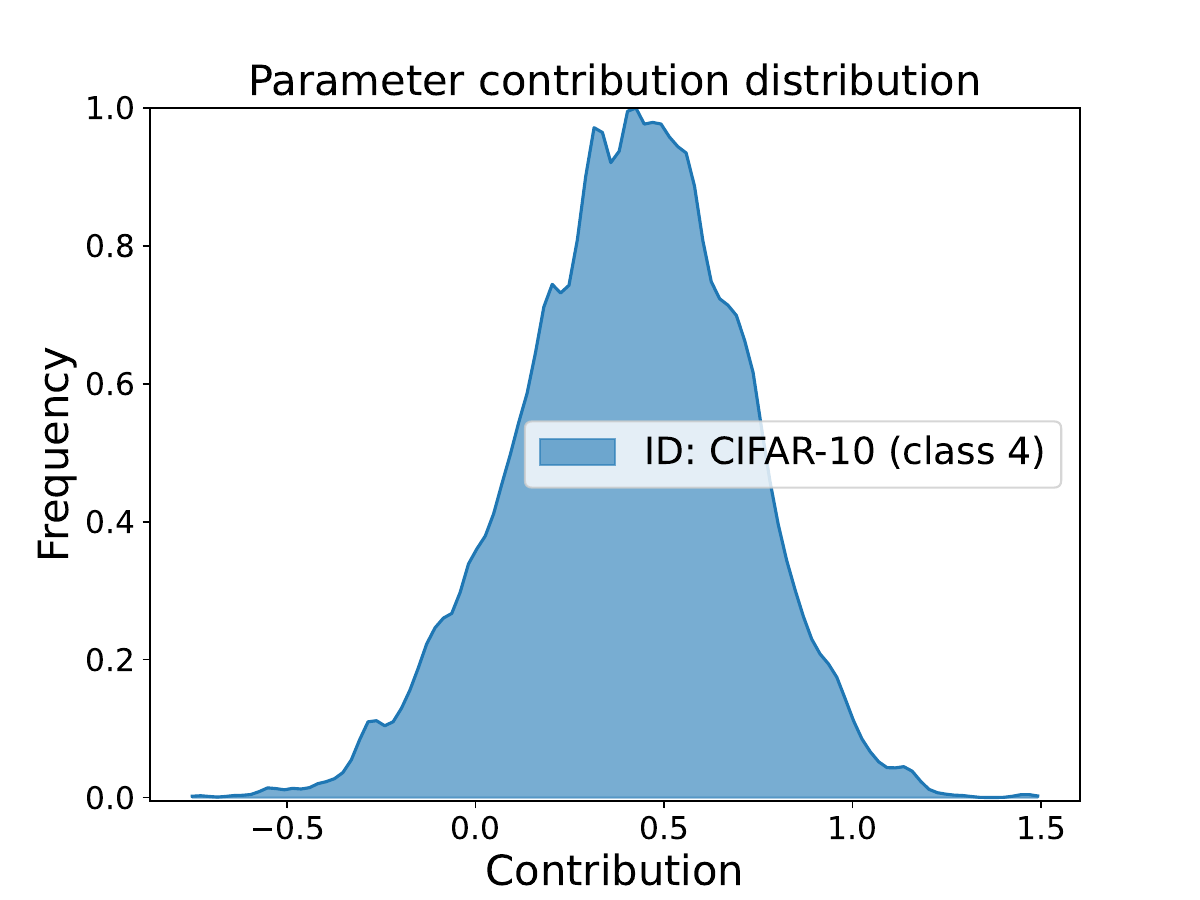}
        \end{subfigure}
    }
    \subcaptionbox{ResNet-18}[1.0\linewidth]{
        \centering
        \begin{subfigure}{0.195\linewidth}
            \centering
            \includegraphics[width=\linewidth]{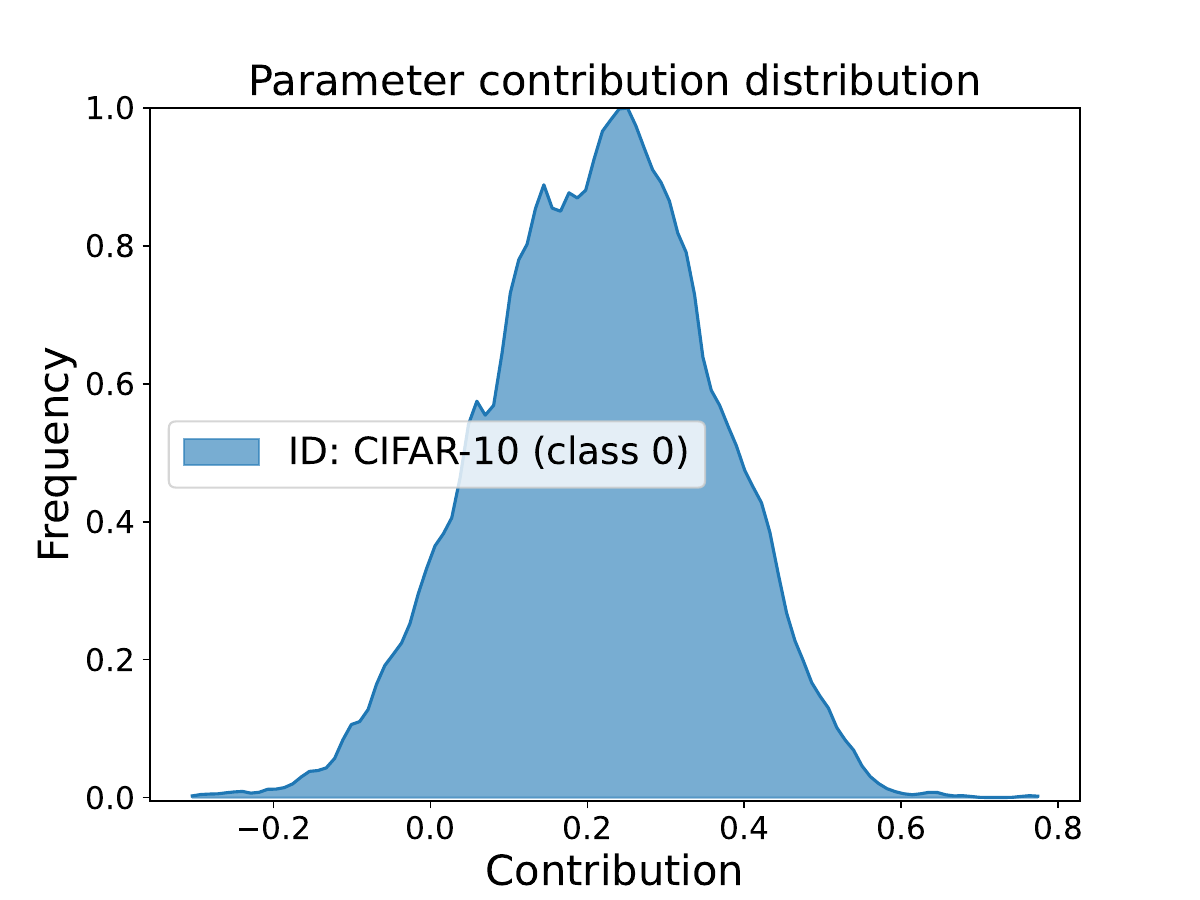}
        \end{subfigure}
        \begin{subfigure}{0.195\linewidth}
            \centering
            \includegraphics[width=\linewidth]{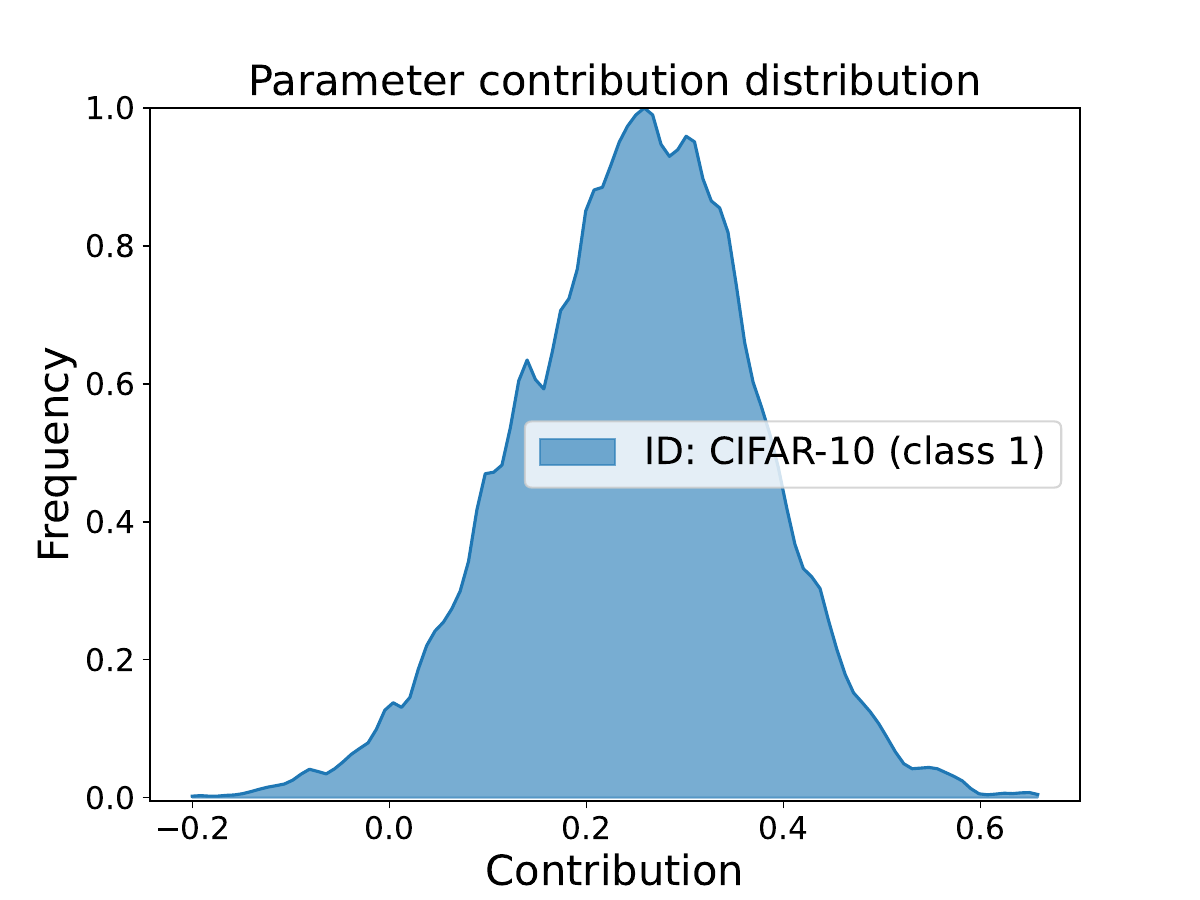}
        \end{subfigure}
        \begin{subfigure}{0.195\linewidth}
            \centering
            \includegraphics[width=\linewidth]{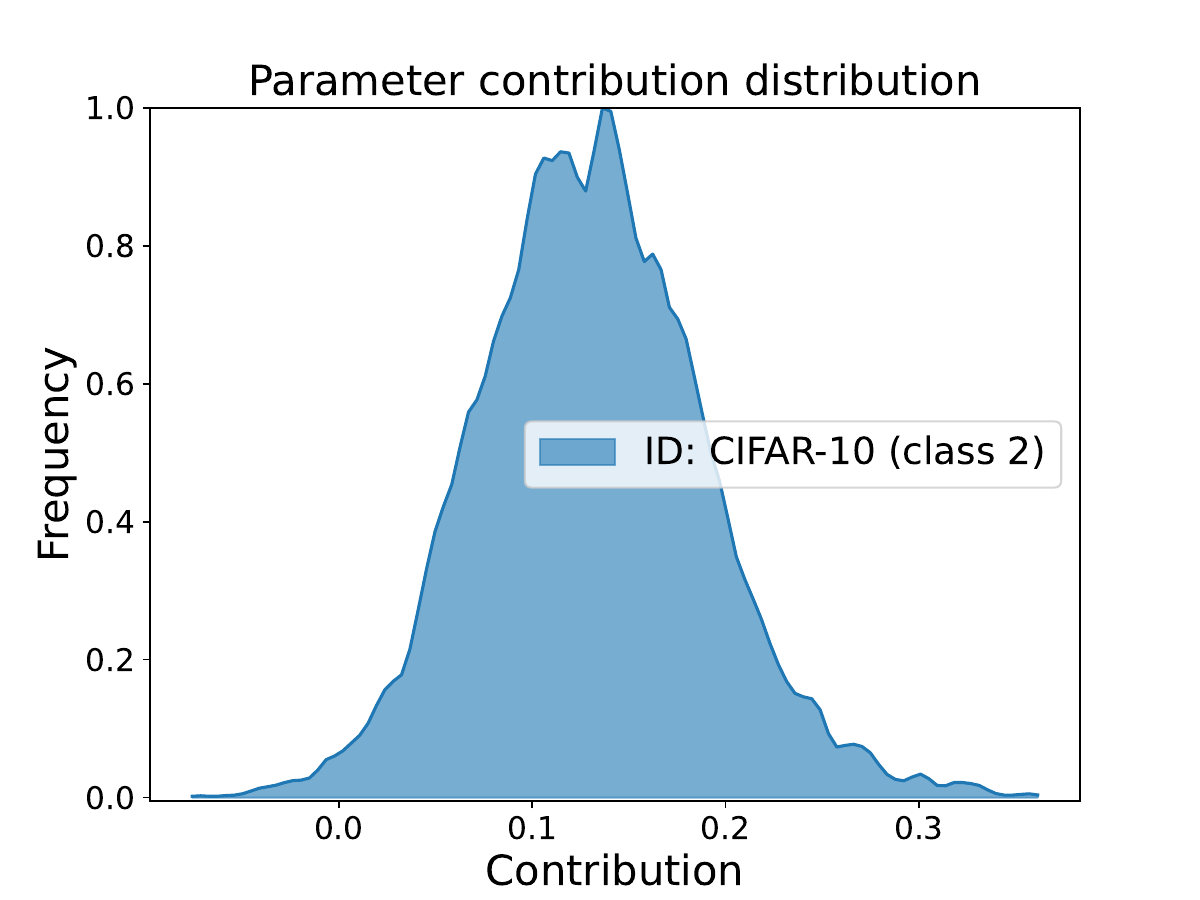}
        \end{subfigure}
        \begin{subfigure}{0.195\linewidth}
            \centering
            \includegraphics[width=\linewidth]{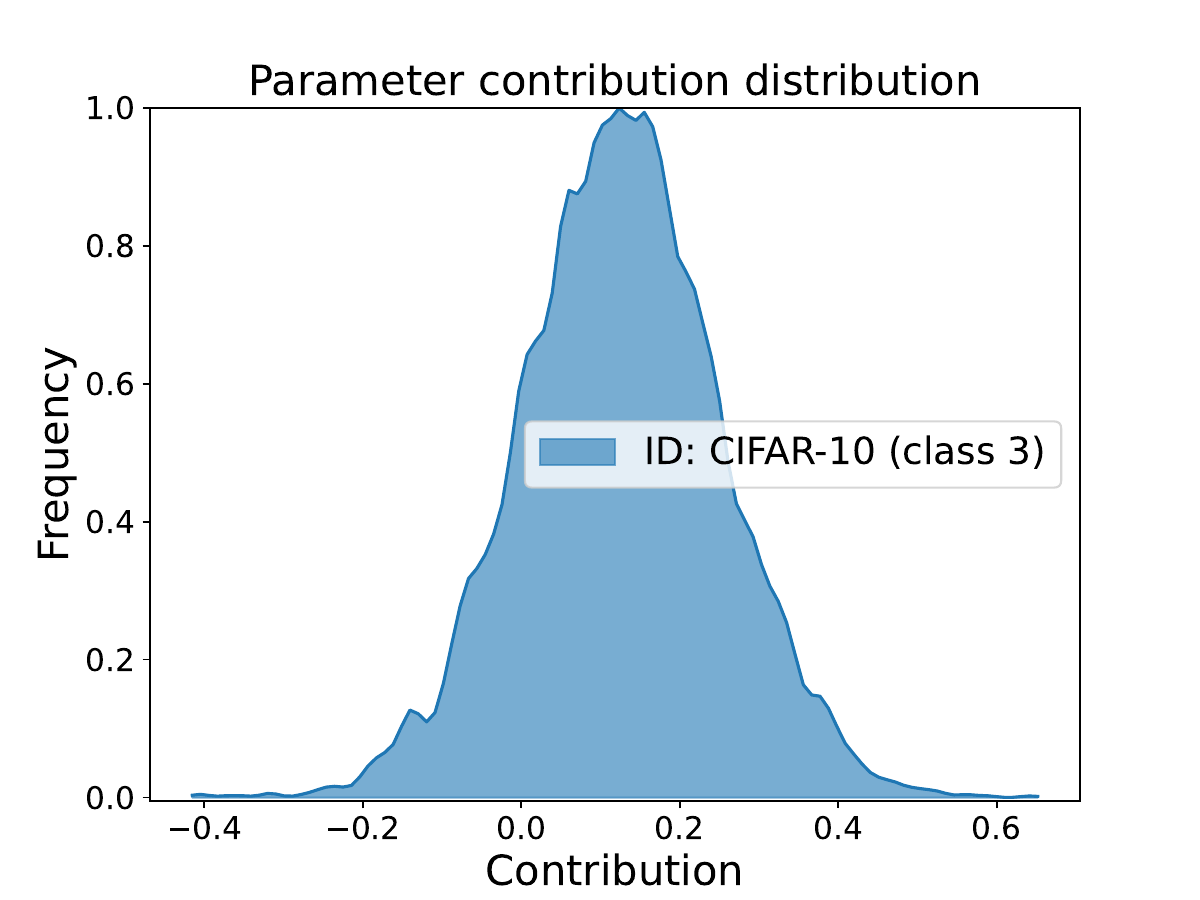}
        \end{subfigure}
        \begin{subfigure}{0.195\linewidth}
            \centering
            \includegraphics[width=\linewidth]{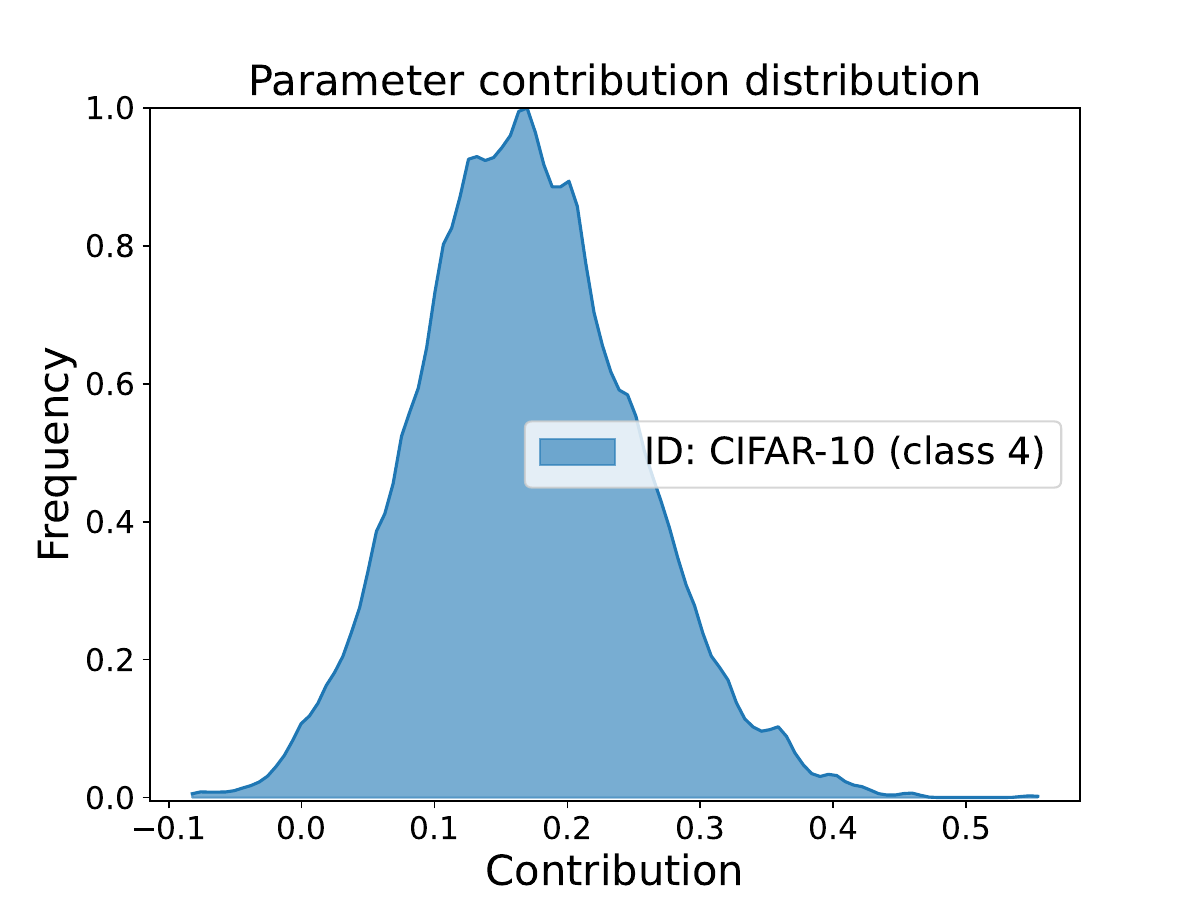}
        \end{subfigure}
    }
    \caption{The distribution of parameter contributions on ID data (CIFAR-10) using (a) DenseNet-101 and (b) ResNet-18 as backbones. These parameters are randomly selected from the weights of the model's last fully connected layer.}
    \label{fig: contribution}
\end{figure*}

Figure \ref{fig: contribution} illustrates the distribution of parameter contributions on ID data. For better visual presentation, we use \mbox{pre-ReLU} activation values to compute the contributions. Empirical evidence indicates that the contributions from the parameters of the last layer approximately follow Gaussian distributions.

\section{Detailed CIFAR Benchmark Results} 
We present the detailed performance results of all six OOD test datasets using DenseNet-101 as the backbone on CIFAR-10 and CIFAR-100 benchmarks in Tables \ref{tab: cifar10} and \ref{tab: cifar100}, respectively. For both tables, weight sparsification methods with ReAct (considering the count of activated features) are grouped separately for a fair comparison.

\section{Software and Hardware} 
All experiments are conducted using Python 3.8.19 and PyTorch 2.0.1, running on an NVIDIA GeForce RTX 4090.

\section{Validation Strategy}
We use a set of Gaussian noise images as the validation set to determine the optimal settings for the redundant parameter pruning percentage $p \in \{5, 10, 15, ..., 50\}$ and the overconfidence parameter pruning threshold $\lambda \in \{1.1, 1.2, 1.3, ..., 3.0\}$. When generating these images, each pixel value is sampled from $\mathcal{N}(0, 1)$. We perform a grid search over all possible values of $p \times \lambda$ and select the parameters that provide the best FPR95 on the validation set as the optimal settings. We adopt the same hyperparameters in the same model in all experiments. For CIFAR-10, the best parameters are $p = 10$ and $\lambda = 2.2$. For CIFAR-100, the optimal settings are $p = 20$ and $\lambda = 1.6$. For ImageNet-1k, $p = 30$ and $\lambda = 1.5$ is optimal.

\section{Details of Pre-trained Models}
For the CIFAR benchmarks, our approach is evaluated using DenseNet-101, with all input images resized to $32 \times 32$ pixels. The models are trained for 100 epochs with batch size 64, weight decay of $1 \times 10^{-4}$, momentum of 0.9, and initial learning rate of 0.1. The learning rate is reduced by a factor of 10 at epochs 50, 75, and 90. The final model achieves accuracies of 94.53\% and 75.09\% on the CIFAR-10 and CIFAR-100 test sets, respectively.

For the ImageNet benchmark, we utilize ResNet-50 as the backbone, employing pre-trained weights obtained from the TorchVision library with an ID accuracy of 76.15\%. All input images are initially resized to $256 \times 256$ pixels, followed by a center crop to $224 \times 224$ pixels.

\section{Comparisons with SOTA methods}
We present a comparative analysis of our proposed method, ITP, against state-of-the-art techniques, namely KNN \cite{DBLP:conf/icml/SunM0L22} and ASH \cite{DBLP:conf/iclr/DjurisicBAL23}, with the corresponding results detailed in Table \ref{tab: sota_cifar} and Table \ref{tab: sota_imagenet}. The experimental findings demonstrate that ITP consistently surpasses the performance of the existing methods across both small-scale and large-scale benchmark datasets.

\section{Comparison on OpenOOD v1.5 Benchmark}
We further provide a standardized evaluation based on the OpenOOD v1.5 large-scale benchmark \cite{DBLP:journals/corr/abs-2306-09301}, comparing the performance in near-OOD and far-OOD scenarios on ImageNet. The results are recorded in Table \ref{tab: openood}, with the baseline performance sourced from the OpenOOD leaderboard.

\section{Relations to Activation-Based Methods}
Many activation-based methods \cite{DBLP:conf/nips/SunGL21,DBLP:conf/iccsip/KongL22,DBLP:conf/nips/ZhuCXLZ00ZC22} have already been proposed to improve OOD detection performance by addressing excessively high activation values. For the recently relevant method LAPS \cite{DBLP:conf/aaai/HeYHWSYLG24}, our work differs in two key aspects. On one hand, LAPS rectifies the activation value to its typical set $[\mu' - \lambda_1\sigma', \mu' + \lambda_2\sigma']$ by pre-computing the mean $\mu'$ and standard deviation $\sigma'$ of activations for each neuron using training data, where $\lambda_1$ and $\lambda_2$ are adaptively adjusted thresholds. Our approach differs in that we focus on identifying more fine-grained anomalies within the weight space. Specifically, during testing, a parameter $\*W_{ij}$ is pruned if its contribution falls within the low-probability range $[\mu_{ij} + \lambda \sigma_{ij}, \infty]$ of its contribution distribution. On the other hand, LAPS treats all classes uniformly for distribution estimation, which introduces interference between classes. In contrast, we address anomalies more precisely by estimating the class-specific distributions of parameter contributions, thereby eliminating potential biases introduced by other classes. The improved performance in our experiments demonstrates the advantages of our approach.

\begin{table*}[ht]
    \renewcommand{\arraystretch}{1.05}
    \caption{OOD detection performance on CIFAR-10 with DenseNet-101 as the backbone. All values are percentages, with the best results in bold. $\uparrow$ indicates that larger values are better, while $\downarrow$ indicates that smaller values are better.}
    \label{tab: cifar10}
    \resizebox{\textwidth}{!}{
        \begin{tabular}{@{\hspace{0.110cm}}c@{\hspace{0.110cm}}c@{\hspace{0.110cm}}c@{\hspace{0.110cm}}c@{\hspace{0.110cm}}c@{\hspace{0.110cm}}c@{\hspace{0.110cm}}c@{\hspace{0.110cm}}c@{\hspace{0.110cm}}c@{\hspace{0.110cm}}c@{\hspace{0.110cm}}c@{\hspace{0.110cm}}c@{\hspace{0.110cm}}c@{\hspace{0.110cm}}c@{\hspace{0.110cm}}c@{\hspace{0.110cm}}}
        \Xhline{1.0pt}
        \multirow{3}{*}{\textbf{Method}} & \multicolumn{12}{c}{\textbf{OOD Datasets}} & \multicolumn{2}{c}{\multirow{2}{*}{\textbf{Average}}} \\
        \cline{2-13}
        & \multicolumn{2}{c}{\textbf{SVHN}} & \multicolumn{2}{c}{\textbf{Textures}} & \multicolumn{2}{c}{\textbf{iSUN}} & \multicolumn{2}{c}{\textbf{LSUN-Resize}} & \multicolumn{2}{c}{\textbf{LSUN-Crop}} & \multicolumn{2}{c}{\textbf{Places365}} & \\
        & FPR95 & AUROC & FPR95 & AUROC & FPR95 & AUROC & FPR95 & AUROC & FPR95 & AUROC & FPR95 & AUROC & FPR95 & AUROC\\
        & $\downarrow$ & $\uparrow$ & $\downarrow$ & $\uparrow$& $\downarrow$ & $\uparrow$& $\downarrow$ & $\uparrow$& $\downarrow$ & $\uparrow$& $\downarrow$ & $\uparrow$& $\downarrow$ & $\uparrow$ \\
        \hline
        MSP & 47.24 & 93.48 & 64.15 & 88.15 & 42.31 & 94.52 & 42.10 & 94.51 & 33.57 & 95.54 & 63.02 & 88.57 & 48.73 & 92.46 \\
        Energy & 40.61 & 93.99 & 56.12 & 86.43 & 10.07 & 98.07 & 9.28 & 98.12 & 3.81 & 99.15 & 39.40 & 91.64 & 26.55 & 94.57\\
        ODIN & 25.29 & 94.57 & 57.50 & 82.38 & \textbf{3.98} & 98.90 & \textbf{3.09} & 99.02 & 4.70 & 98.86 & 52.85 & 88.55 & 24.57 & 93.71 \\
        ReAct & 41.64 & 93.87 & 43.58 & 92.47 & 12.72 & 97.72 & 11.46 & 97.87 & 5.96 & 98.84 & 43.31 & 91.03 & 26.45 & 94.67 \\ 
        DICE& 25.99 & 95.90 & 41.90 & 88.18 & 4.36 & \textbf{99.14} & 3.91 & \textbf{99.20} & \textbf{0.26} & \textbf{99.92} & 48.59 & 89.13 & 20.83 & 95.24\\
        OPNP & 26.97 & 95.73 & 44.75 & 87.96 & 8.43 & 98.21 & 7.60 & 98.26 & 3.66 & 99.16 & 41.02 & 91.54 & 22.07 & 95.14  \\
        LAPS & 21.83 & 96.21 & 37.80 & 92.38 & 7.81 & 98.29 & 6.41 & 98.38 & 3.44 & 99.17 & \textbf{39.11} & \textbf{92.14} & 19.40 & 96.10 \\
        \textbf{ITP (Ours)} & \textbf{15.96} & \textbf{97.10} & \textbf{28.21} & \textbf{94.66} & 5.45 & 98.99 & 4.46 & 99.05 & 0.47 & 99.86 & 45.75 & 90.17 & \textbf{16.72} & \textbf{96.64}\\
        \Xhline{0.5pt}
        DICE + ReAct& 12.49 & 97.61 & 25.83 & 94.56 & 5.27 & 99.02 & 3.95 & 99.14 & \textbf{0.43} & \textbf{99.89} & 50.94 & 89.63 & 16.48 & 96.64\\
        OPNP + ReAct & 21.21 & 96.43 & 34.66 & 92.85 & 6.31 & 98.57 & 6.26 & 98.62 & 3.60 & 99.20 & \textbf{38.71} & \textbf{92.44} & 18.46 & 96.35  \\
        LINe (w/ ReAct)& \textbf{11.38} & \textbf{97.75} & 23.44 & 95.12 & 4.90 & 99.01 & 4.19 & 99.09 & 0.61 & 99.83 & 43.78 & 91.12 & 14.72 & 96.99\\
        \textbf{ITP + ReAct(Ours)} & 11.87 & 97.70 & \textbf{21.72} & \textbf{95.78} & \textbf{4.43} & \textbf{99.16} & \textbf{3.31} & \textbf{99.27} & 0.53 & 99.86 & 45.16 & 90.98 & \textbf{14.50} & \textbf{97.13}\\
        \Xhline{1.0pt}
        \end{tabular}
    }
\end{table*}

\begin{table*}[ht]
    \renewcommand{\arraystretch}{1.05}
    \caption{OOD detection performance on CIFAR-100 with DenseNet-101 as the backbone.}
    \label{tab: cifar100}
    \resizebox{\textwidth}{!}{
        \begin{tabular}{@{\hspace{0.10cm}}c@{\hspace{0.10cm}}c@{\hspace{0.10cm}}c@{\hspace{0.10cm}}c@{\hspace{0.10cm}}c@{\hspace{0.10cm}}c@{\hspace{0.10cm}}c@{\hspace{0.10cm}}c@{\hspace{0.10cm}}c@{\hspace{0.10cm}}c@{\hspace{0.10cm}}c@{\hspace{0.10cm}}c@{\hspace{0.10cm}}c@{\hspace{0.10cm}}c@{\hspace{0.10cm}}c@{\hspace{0.10cm}}}
        \Xhline{1.0pt}
        \multirow{3}{*}{\textbf{Method}} & \multicolumn{12}{c}{\textbf{OOD Datasets}} & \multicolumn{2}{c}{\multirow{2}{*}{\textbf{Average}}} \\
        \cline{2-13}
        & \multicolumn{2}{c}{\textbf{SVHN}} & \multicolumn{2}{c}{\textbf{Textures}} & \multicolumn{2}{c}{\textbf{iSUN}} & \multicolumn{2}{c}{\textbf{LSUN-Resize}} & \multicolumn{2}{c}{\textbf{LSUN-Crop}} & \multicolumn{2}{c}{\textbf{Places365}} & \\
        & FPR95 & AUROC & FPR95 & AUROC & FPR95 & AUROC & FPR95 & AUROC & FPR95 & AUROC & FPR95 & AUROC & FPR95 & AUROC\\
        & $\downarrow$ & $\uparrow$ & $\downarrow$ & $\uparrow$& $\downarrow$ & $\uparrow$& $\downarrow$ & $\uparrow$& $\downarrow$ & $\uparrow$& $\downarrow$ & $\uparrow$& $\downarrow$ & $\uparrow$ \\
        \Xhline{0.5pt}
        MSP & 81.70 & 75.40 & 84.79 & 71.48 & 85.99 & 70.17 & 85.24 & 69.18 & 60.49 & 85.60 & 82.55 & 74.31 & 80.13 & 74.36 \\
        Energy& 87.46 & 81.85 & 84.15 & 71.03 & 74.54 & 78.95 &  70.65 & 80.14 &  14.72 & 97.43 & \textbf{79.20} & \textbf{77.72} & 68.45 & 81.19\\
        ODIN & 41.35 & 92.65 & 82.34 & 71.48 & 67.05 & 83.84 & 65.22 & 84.22 & 10.54 & 97.93 & 82.32 & 76.84 & 58.14 & 84.49\\
        ReAct& 83.81 & 81.41 & 77.78 & 78.95 & 65.27 & 86.55 & 60.08 & 87.88 & 25.55 & 94.92 &  82.65 & 74.04 & 62.27 & 84.47\\
        DICE & 54.65 & 88.84 & 65.04 & 76.42 & 48.72 & 90.08 & 49.40 & 91.04 & \textbf{0.93} & \textbf{99.74} & 79.58 & 77.26 & 49.72 & 87.23 \\
        OPNP & 57.02 & 87.10 & 70.96 & 77.37 & 50.02 & 90.91 & 46.10 & 92.04 & 6.40 & 98.57 & 80.21 & 77.19 & 51.79 & 87.20  \\
        LAPS & 52.25 & 88.72 & 61.29 & 85.04 & 48.87 & 90.23 & 45.79 & 92.43 & 14.99 & 96.58 & 79.81 & 75.42 & 50.50 & 88.07 \\
        \textbf{ITP (Ours)} & \textbf{28.15} & \textbf{94.58} & \textbf{32.45} & \textbf{92.06} & \textbf{29.88} & \textbf{94.16} & \textbf{35.51} & \textbf{93.16} & 1.82 & 99.54 & 82.39 & 74.83 & \textbf{35.03} & \textbf{91.39}\\
        \Xhline{0.5pt}
        DICE + ReAct& 55.52 & 88.02 & 41.54 & 86.26 & 44.32 & 91.44 & 54.44 & 89.84 & 7.56 & 98.61 & 94.05 & 56.26 & 49.57 & 85.07\\
        OPNP + ReAct & 52.57 & 88.61 & 55.48 & 84.61 & 29.20 & 93.14 & 26.50 & 93.77 & 11.89 & 97.13 & \textbf{82.26} & \textbf{74.02} & 42.98 & 88.55  \\
        LINe (w/ ReAct) & 31.10 & 91.90 & 39.29 & 87.84 & 24.07 & 94.85 & 25.32 & 94.63 & 5.72 & 98.87 & 88.50 & 63.93 & 35.67 & 88.67 \\
        \textbf{ITP + ReAct(Ours)} & \textbf{26.43} & \textbf{94.00} & 
        \textbf{27.82} & \textbf{92.40} & \textbf{17.15} & \textbf{96.83} & \textbf{21.32} & \textbf{96.20} & \textbf{4.38} & \textbf{99.10} & 83.69 & 72.96 & \textbf{30.13} & \textbf{91.91} \\
        \Xhline{1.0pt}
        \end{tabular}
    }
\end{table*}

\clearpage

\begin{table*}[ht]
    \renewcommand{\arraystretch}{1.05}
    \caption{Comparisons with SOTA methods on CIFAR benckmarks.}
    \label{tab: sota_cifar}
    \resizebox{\textwidth}{!}{
        \begin{tabular}{@{\hspace{0.10cm}}c@{\hspace{0.10cm}}c@{\hspace{0.10cm}}c@{\hspace{0.10cm}}c@{\hspace{0.10cm}}c@{\hspace{0.10cm}}c@{\hspace{0.10cm}}c@{\hspace{0.10cm}}c@{\hspace{0.10cm}}c@{\hspace{0.10cm}}c@{\hspace{0.10cm}}c@{\hspace{0.10cm}}c@{\hspace{0.10cm}}c@{\hspace{0.10cm}}c@{\hspace{0.10cm}}c@{\hspace{0.10cm}}}
        \Xhline{1.0pt}
        \multirow{3}{*}{\textbf{Method}} & \multicolumn{12}{c}{\textbf{OOD Datasets}} & \multicolumn{2}{c}{\multirow{2}{*}{\textbf{Average}}} \\
        \cline{2-13}
        & \multicolumn{2}{c}{\textbf{SVHN}} & \multicolumn{2}{c}{\textbf{Textures}} & \multicolumn{2}{c}{\textbf{iSUN}} & \multicolumn{2}{c}{\textbf{LSUN-Resize}} & \multicolumn{2}{c}{\textbf{LSUN-Crop}} & \multicolumn{2}{c}{\textbf{Places365}} & \\
        & FPR95 & AUROC & FPR95 & AUROC & FPR95 & AUROC & FPR95 & AUROC & FPR95 & AUROC & FPR95 & AUROC & FPR95 & AUROC\\
        & $\downarrow$ & $\uparrow$ & $\downarrow$ & $\uparrow$& $\downarrow$ & $\uparrow$& $\downarrow$ & $\uparrow$& $\downarrow$ & $\uparrow$& $\downarrow$ & $\uparrow$& $\downarrow$ & $\uparrow$ \\
        \Xhline{0.5pt}
        \multicolumn{15}{c}{\textit{CIFAR-10 benckmark}} \\
        KNN & \textbf{3.96} & \textbf{99.29} & \textbf{19.56} & \textbf{96.40} & 10.20 & 98.21 & 9.90 & 98.12 & 6.93 & 98.75 & 46.84 & 90.06 & 16.23 & 96.81 \\
        ASH-P & 30.14 & 95.29 & 50.85 & 88.29 & 8.46 & 98.29 & 7.97 & 98.33 & 2.82 & 99.34 & \textbf{40.46} & \textbf{91.76} & 23.45 & 95.22\\
        ASH-B & 17.92 & 96.86 & 35.73 & 92.88 & 8.59 & 98.45 & 8.13 & 98.54 & 2.52 & 99.48 & 48.47 & 89.93 & 20.23 & 96.02 \\
        ASH-S & 6.51 & 98.65 & 24.34 & 95.05 & 5.17 & 98.90 & 4.96 & 98.92 & 0.90 & 99.73 & 48.45 & 88.34 & 15.05 & 96.61 \\
        \textbf{ITP (Ours)} & 15.96 & 97.10 & 28.21 & 94.66 & 5.45 & 98.99 & 4.46 & 99.05 & \textbf{0.47} & \textbf{99.86} & 45.75 & 90.17 & 16.72 & 96.64\\
        \textbf{ITP + ReAct(Ours)} & 11.87 & 97.70 & 21.72 & 95.78 & \textbf{4.43} & \textbf{99.16} & \textbf{3.31} & \textbf{99.27} & 0.53 & \textbf{99.86} & 45.16 & 90.98 & \textbf{14.50} & \textbf{97.13}\\
        \Xhline{0.5pt}
        \multicolumn{15}{c}{\textit{CIFAR-100 benckmark}} \\
        KNN & \textbf{19.50} & \textbf{96.07} & 24.75 & 93.40 & 36.57 & 92.73 & 43.79 & 91.56 & 32.76 & 92.79 & 93.83 & 58.56 & 41.87 & 87.52\\
        ASH-P & 81.86 & 83.86 & \textbf{11.60} & \textbf{97.89} & 67.56 & 81.67 & 70.90 & 80.81 & 78.24 & 74.09 & \textbf{77.03} & \textbf{77.94} & 64.53 & 82.71\\
        ASH-B & 53.52 & 90.27 & 53.71 & 84.25 & 47.82 & 91.09 & 48.38 & 91.03 & 4.46 & 99.17 & 84.52 & 72.46 & 48.73 & 88.04 \\
        ASH-S & 25.05 & 95.76 & 34.02 & 92.35 & 46.67 & 91.30 & 51.33 & 90.12 & 5.52 & 98.94 & 85.86 & 71.62 & 41.40 & 90.02 \\
        
        \textbf{ITP (Ours)} & 28.15 & 94.58 & 32.45 & 92.06 & 29.88 & 94.16 & 35.51 & 93.16 & \textbf{1.82} & \textbf{99.54} & 82.39 & 74.83 & 35.03 & 91.39 \\
        \textbf{ITP + ReAct(Ours)} & 26.43 & 94.00 & 27.82 & 92.40 & \textbf{17.15} & \textbf{96.83} & \textbf{21.32} & \textbf{96.20} & 4.38 & 99.10 & 83.69 & 72.96 & \textbf{30.13} & \textbf{91.91} \\
        \Xhline{1.0pt}
        \end{tabular}
    }
\end{table*}

\begin{table*}[t]
    \renewcommand{\arraystretch}{1.05}
    \caption{Comparisons with SOTA methods on ImageNet benckmark.}
    \centering
    \label{tab: sota_imagenet}
    \begin{tabular}{@{\hspace{0.30cm}}c@{\hspace{0.30cm}}c@{\hspace{0.30cm}}c@{\hspace{0.30cm}}c@{\hspace{0.30cm}}c@{\hspace{0.30cm}}c@{\hspace{0.30cm}}c@{\hspace{0.30cm}}c@{\hspace{0.30cm}}c@{\hspace{0.30cm}}c@{\hspace{0.30cm}}c@{\hspace{0.30cm}}}
    \Xhline{1.0pt}
     \multirow{4}{*}{\textbf{Method}} & \multicolumn{8}{c}{\textbf{OOD Datasets}} & \multicolumn{2}{c}{\multirow{2}{*}{\textbf{Average}}} \\
    \cline{2-9}
    & \multicolumn{2}{c}{\textbf{iNaturalist}} & \multicolumn{2}{c}{\textbf{SUN}} & \multicolumn{2}{c}{\textbf{Places}} & \multicolumn{2}{c}{\textbf{Textures}} \\
    & FPR95 & AUROC & FPR95 & AUROC & FPR95 & AUROC & FPR95 & AUROC & FPR95 & AUROC \\
    & $\downarrow$ & $\uparrow$& $\downarrow$ & $\uparrow$& $\downarrow$ & $\uparrow$& $\downarrow$ & $\uparrow$& $\downarrow$ & $\uparrow$ \\
    \Xhline{0.5pt}
    KNN & 59.77 & 85.89 & 68.88 & 80.08 & 78.15 & 74.10 & 10.90 & 97.42 & 54.68 & 84.37\\
    ASH-P & 44.57 & 92.51 & 52.88 & 88.35 & 61.79 & 85.58 & 42.06 & 89.70 & 50.32 & 89.04 \\
    ASH-B & 14.21 & 97.32 & \textbf{22.08} & \textbf{95.10} & 33.45 & \textbf{92.31} & 21.17 & 95.50 & 22.73 & 95.06 \\
    ASH-S & 11.49 & 97.87 & 27.98 & 94.02 & 39.78 & 90.98 & \textbf{11.93} & \textbf{97.60} & 22.80 & 95.12 \\
    \textbf{ITP (Ours)} & 11.53 & 97.83 & 25.82 & 93.58 & 35.63 & 90.75 & 17.06 & 96.03 & 22.51 & 94.55 \\
    \textbf{ITP + ReAct (Ours)}  & \textbf{9.78} & \textbf{98.02} & 22.82 & 94.47 & \textbf{30.87} & 92.03 & 18.09 & 95.98 & \textbf{20.39} & \textbf{95.13} \\
    \Xhline{1.0pt}
    \end{tabular}
\end{table*}

\begin{table*}[t]
    \renewcommand{\arraystretch}{1.05}
    \caption{Performance comparison (FPR95$\downarrow$/AUROC$\uparrow$) on OpenOOD v1.5 benchmark.}
    \centering
    \label{tab: openood}
    \begin{tabular}{@{\hspace{0.23cm}}c|@{\hspace{0.23cm}}c@{\hspace{0.23cm}}c@{\hspace{0.23cm}}c|@{\hspace{0.23cm}}c@{\hspace{0.23cm}}c@{\hspace{0.23cm}}c@{\hspace{0.23cm}}c@{\hspace{0.23cm}}}
    \Xhline{1.0pt}
    \multirow{1}{*}{\textbf{ImageNet-1k}} & \multicolumn{3}{c}{\textbf{Near OOD}} & \multicolumn{4}{c}{\textbf{Far OOD}} \\ 
    \multirow{1}{*}{\textbf{ResNet-50}} &  SSB-hard & NINCO & Average & iNaturalist & Textures & OpenImage-O &  Average \\
    \Xhline{0.5pt}
    MSP & 74.49 / 72.09 & 56.88 / 79.97 & 65.68 / 76.02 & 43.34 / 88.41 & 60.87 / 82.43 & 50.13 / 84.86 & 51.45 / 85.23 \\
    ODIN & 76.83 / 71.74 & 68.16 / 77.77 & 72.50 / 74.75 & 35.98 / 91.17 & 49.24 / 89.00 & 46.67 / 88.23 & 43.96 / 89.47 \\
    KNN & 83.36 / 62.57 & 58.39 / 79.64 & 70.87 / 71.10 & 40.80 / 86.41 & 17.31 / 97.09 & 44.27 / 87.04 & 34.13 / 90.18 \\
    ASH & 73.66 / 72.89 & 52.97 / 83.45 & 63.32 / 78.17 & 14.04 / 97.07 & 15.26 / 96.90 & 29.15 / 93.26 & 19.49 / 95.74  \\
    Energy & 76.54 / 72.08 & 60.59 / 79.70 & 68.56 / 75.89 & 31.33 / 90.63 & 45.77 / 88.70 & 38.08 / 89.06 & 38.40 / 89.47 \\
    DICE & 77.96 / 70.13  & 66.90 / 76.01 & 72.43 / 73.07 & 33.37 / 92.54 & 44.28 / 92.04 & 47.83 / 88.26 & 41.83 / 90.25 \\
    \textbf{ITP} & 76.13 / 73.33 & 58.90 / 81.43 & 67.52 / 77.38 & 17.85 / 96.31 & 28.10 / 94.51 & 36.32 / 91.67 & 27.42 / 94.17\\
    \textbf{ITP + ReAct} & 76.72 / 72.86 & 54.83 / 82.34 & 65.77 / 77.60 & 14.71 / 96.82 & 22.63 / 95.11 & 32.44 / 92.43 & 23.26 / 94.79 \\
    \Xhline{1.0pt}
    \end{tabular}
\end{table*}
\end{document}